\PassOptionsToPackage{table}{xcolor}
\PassOptionsToPackage{a4paper}{geometry}
\documentclass{style}
\usepackage[T1]{fontenc}
\usepackage{amsmath,amssymb}
\usepackage{graphicx}
\usepackage{booktabs}
\usepackage{multirow}
\usepackage{array}
\usepackage{tabularx}
\usepackage{xltabular}  %
\usepackage{adjustbox}
\usepackage{listings}
\usepackage{enumitem}
\usepackage{needspace}
\usepackage{microtype}
\usepackage{makecell}
\usepackage[scaled=0.92]{zi4}
\usepackage{algorithm}
\usepackage[noend]{algpseudocode}
\usepackage{xspace}
\usepackage{csquotes}
\usepackage{float}       %

\usepackage[osf=false]{newpxtext}
\usepackage[varbb]{newpxmath}
\usepackage{pifont}   %
\definecolor{rsired}{HTML}{8B0000}      %
\definecolor{rsiredbg}{HTML}{F7EDED}
\definecolor{rsislate}{HTML}{596675}    %
\definecolor{rsislatebg}{HTML}{E7ECF1}
\definecolor{rsiblue}{HTML}{0072B2}     %

\newcounter{finding}
\providecommand{\findingname}{Finding}
\newtcolorbox{finding}{enhanced, breakable=false, colback=rsiredbg, colframe=rsiredbg, boxrule=0pt,
  borderline west={2pt}{0pt}{rsired}, arc=0pt, outer arc=0pt, left=8pt, right=8pt, top=5pt, bottom=5pt,
  before skip=8pt, after skip=8pt, fontupper=\itshape,
  before upper={\stepcounter{finding}{\upshape\bfseries\color{rsired}\findingname~\thefinding.}~}}

\newtcolorbox{questions}[1][Research questions]{enhanced, colback=rsislatebg!55, colframe=rsislate!70,
  boxrule=0.4pt, arc=1.5mm, left=8pt, right=8pt, top=4pt, bottom=5pt, before skip=6pt, after skip=10pt,
  title={\sffamily\bfseries #1}, coltitle=white, colbacktitle=rsislate, toptitle=2pt, bottomtitle=2pt}

\newcommand{\claim}[1]{\textbf{#1}\hspace{0.4em}}

\colorlet{gaintint}{rsired!6}

\makeatletter
\renewcommand{\BreakableUnderscore}{\leavevmode\prw@zbreak\textunderscore\discretionary{}{}{}\prw@zbreak}
\makeatother
\expandafter\def\csname pv@AUDIT_TASKS\endcsname{42}
\expandafter\def\csname pv@AUDIT_CL_TASKS\endcsname{20}
\expandafter\def\csname pv@AUDIT_CL_VARIANTS\endcsname{26}
\expandafter\def\csname pv@AUDIT_HIDDEN\endcsname{13}
\expandafter\def\csname pv@AUDIT_THRESH\endcsname{13}
\expandafter\def\csname pv@AUDIT_EVENT\endcsname{10}
\expandafter\def\csname pv@AUDIT_DEFECT\endcsname{9}
\expandafter\def\csname pv@AUDIT_ENDSTATE\endcsname{5}
\expandafter\def\csname pv@AUDIT_LOOSE\endcsname{4}
\expandafter\def\csname pv@AUDIT_CONTRA\endcsname{3}
\expandafter\def\csname pv@AUDIT_HOME_TASKS\endcsname{33}
\expandafter\def\csname pv@AUDIT_HOME_INSTR\endcsname{5}
\expandafter\def\csname pv@LEARN_N\endcsname{25}
\expandafter\def\csname pv@LEARN_CL\endcsname{16}
\expandafter\def\csname pv@LEARN_FAITH\endcsname{9}
\expandafter\def\csname pv@T_ALL_V0\endcsname{24.6}
\expandafter\def\csname pv@T_ALL_OFF\endcsname{47.8}
\expandafter\def\csname pv@T_ALL_ON\endcsname{62.0}
\expandafter\def\csname pv@T_FA_V0\endcsname{53.3}
\expandafter\def\csname pv@T_FA_OFF\endcsname{83.3}
\expandafter\def\csname pv@T_FA_ON\endcsname{80.6}
\expandafter\def\csname pv@T_CL_V0\endcsname{8.4}
\expandafter\def\csname pv@T_CL_OFF\endcsname{27.8}
\expandafter\def\csname pv@T_CL_ON\endcsname{51.6}
\expandafter\def\csname pv@RJ16_ORIG_ONONLY\endcsname{93}
\expandafter\def\csname pv@RJ16_ORIG_OFFONLY\endcsname{17}
\expandafter\def\csname pv@RJ16_FAITH_ONONLY\endcsname{47}
\expandafter\def\csname pv@RJ16_FAITH_OFFONLY\endcsname{54}
\expandafter\def\csname pv@RJ16_GAPCHG\endcsname{25.9}
\expandafter\def\csname pv@RJ16_GAPCHG_LO\endcsname{4.7}
\expandafter\def\csname pv@RJ16_GAPCHG_HI\endcsname{54.4}
\expandafter\def\csname pv@SW_TEST_ORIG\endcsname{13.5}
\expandafter\def\csname pv@SW_TEST_FAITH\endcsname{65.0}
\expandafter\def\csname pv@SW_ON_ORIG\endcsname{57.0}
\expandafter\def\csname pv@SW_ON_FAITH\endcsname{45.0}
\expandafter\def\csname pv@CARDS_ON_N\endcsname{55}
\expandafter\def\csname pv@CARDS_ON_CRIT\endcsname{19}
\expandafter\def\csname pv@CARDS_ON_CHARS\endcsname{47}
\expandafter\def\csname pv@CARDS_OFF_N\endcsname{53}
\expandafter\def\csname pv@CARDS_DROPPED\endcsname{9}
\expandafter\def\csname pv@ON_GAIN\endcsname{14.2}
\expandafter\def\csname pv@LB_NOTFULL\endcsname{6}
\expandafter\def\csname pv@LB_TASKS\endcsname{40}
\expandafter\def\csname pv@CS_OPEN_V0\endcsname{35}
\expandafter\def\csname pv@CS_OPEN_OFF\endcsname{91}
\expandafter\def\csname pv@CS_OPEN_ON\endcsname{3}
\expandafter\def\csname pv@CS_EMPTY_HALF\endcsname{28}
\expandafter\def\csname pv@CS_EMPTY_OPEN\endcsname{3}
\expandafter\def\csname pv@CS_SOLVE_V0\endcsname{14}
\expandafter\def\csname pv@CS_SOLVE_OFF\endcsname{19}
\expandafter\def\csname pv@CS_SOLVE_ON\endcsname{15}
\expandafter\def\csname pv@CS_TUBES_V0\endcsname{12}
\expandafter\def\csname pv@CS_TUBES_OFF\endcsname{11}
\expandafter\def\csname pv@CS_TUBES_ON\endcsname{6}
\expandafter\def\csname pv@CS_HOVER_ON\endcsname{116}
\expandafter\def\csname pv@CS_HOVER_V0\endcsname{40}
\expandafter\def\csname pv@CS_HOVER_OFF\endcsname{57}
\expandafter\def\csname pv@RCA_ATOMIC\endcsname{55}
\expandafter\def\csname pv@RCA_ATOMIC_PAPER\endcsname{59}
\expandafter\def\csname pv@RCC\endcsname{54}
\expandafter\def\csname pv@RCC_PAPER\endcsname{32}
\expandafter\def\csname pv@T_ALL_OFF_D\endcsname{+23.2}
\expandafter\def\csname pv@T_ALL_ON_D\endcsname{+37.4}
\expandafter\def\csname pv@T_FA_OFF_D\endcsname{+30.0}
\expandafter\def\csname pv@T_FA_ON_D\endcsname{+27.3}
\expandafter\def\csname pv@T_CL_OFF_D\endcsname{+19.4}
\expandafter\def\csname pv@T_CL_ON_D\endcsname{+43.2}
\expandafter\def\csname pv@XB_USE_R10_OFF\endcsname{5}
\expandafter\def\csname pv@XB_USE_R10_ON\endcsname{20.5}
\expandafter\def\csname pv@XB_LB_ALL_V0\endcsname{188}
\expandafter\def\csname pv@RJ16_ON_ORIG\endcsname{51.6}
\expandafter\def\csname pv@RJ16_ON_FAITH\endcsname{45.0}
\expandafter\def\csname pv@RJ16_OFF_ORIG\endcsname{27.8}
\expandafter\def\csname pv@RJ16_OFF_FAITH\endcsname{47.2}
\expandafter\def\csname pv@LB_FULL\endcsname{34}
\expandafter\def\csname pv@XB_LB_CALL_OFF\endcsname{5}
\expandafter\def\csname pv@XB_LB_CALL_ON\endcsname{12}
\expandafter\def\csname pv@XB_RC_CALL_RC\endcsname{158}
\expandafter\def\csname pv@XB_LB_T5\endcsname{6 / 8 / 5}
\expandafter\def\csname pv@XB_LB_T9\endcsname{2 / 3 / 3}
\expandafter\def\csname pv@XB_LB_HALF_V0\endcsname{21 / 24}
\expandafter\def\csname pv@XB_LB_HALF_OFF\endcsname{22 / 26}
\expandafter\def\csname pv@XB_LB_HALF_ON\endcsname{25 / 21}
\expandafter\def\csname pv@TEST_TASKS_N\endcsname{20}
\expandafter\def\csname pv@HO_ALL_EPS\endcsname{500}
\expandafter\def\csname pv@RERUN_DIFFER_PCT\endcsname{4}
\expandafter\def\csname pv@R0_CROSS_DIFFER_PCT\endcsname{13}

\makeatletter
\newcommand{\V}[1]{\ifcsname pv@#1\endcsname\csname pv@#1\endcsname\else
  \PackageError{values}{Unknown value #1}{Add it to generated/values.tex}\fi}
\makeatother

\newcommand{\wCE}{w/~CE\xspace}
\newcommand{\woCE}{w/o~CE\xspace}
\newcommand{\Corig}{C^{\text{orig}}}
\newcommand{\Cfaith}{C^{\text{faith}}}
\newcommand{\SRo}{\text{SR}^{\text{orig}}}
\newcommand{\SRf}{\text{SR}^{\text{faith}}}
\newcommand{\Lempty}{\mathcal{L}_\varnothing}
\newcommand{\LCEo}{\mathcal{L}_{\text{CE}}^{\text{orig}}}
\newcommand{\LCEf}{\mathcal{L}_{\text{CE}}^{\text{faith}}}

\newcommand{\task}[1]{\texttt{#1}}

\definecolor{codebg}{HTML}{F7F6F2}
\definecolor{coderule}{HTML}{D9DCD2}
\lstdefinestyle{card}{basicstyle=\ttfamily\scriptsize, columns=fullflexible, keepspaces=true, breaklines=true,
  backgroundcolor=\color{codebg}, frame=single, rulecolor=\color{coderule}, framesep=3pt, xleftmargin=3pt,
  xrightmargin=3pt, aboveskip=2pt, belowskip=0pt}
\setlist{leftmargin=*,itemsep=1pt,topsep=2pt,parsep=0pt}
\colorlet{uptint}{rsired!8}
\colorlet{downtint}{rsislate!14}

\definecolor{gaingreen}{HTML}{2E7D32}
\definecolor{lossred}{HTML}{B23A3A}
\newcommand{\dpos}[1]{\textcolor{gaingreen}{(+#1)}}

\title{Learning the Task or Learning the Verifier?\\ Rethinking Criteria Exposure in Embodied Self-Improvement}

\renewcommand{\authorfont}{\fontsize{11.5}{14}\selectfont}
\author{
  Wenbo~Huang\texorpdfstring{$^{1}$}{} \hspace{0.1cm}
  Jianxin~Bi\texorpdfstring{$^{1,2}$}{} \hspace{0.1cm}
  Shunyu~Yang\texorpdfstring{$^{1}$}{} \hspace{0.1cm}
  Ruiyang~Si\texorpdfstring{$^{1}$}{}\\
  Rui~Ni\texorpdfstring{$^{1}$}{} \hspace{0.1cm}
  Qiang~Wang\texorpdfstring{$^{1}$}{} \hspace{0.1cm}
  Shulong~Jiang\texorpdfstring{$^{1}$}{} \hspace{0.1cm}
  Yintianrun~Zhang\texorpdfstring{$^{1}$}{}\\
  Feixiang~Ruan\texorpdfstring{$^{1}$}{} \hspace{0.1cm}
  Kewei~Zhang\texorpdfstring{$^{1}$}{} \hspace{0.1cm}
  Ye~Huang\texorpdfstring{$^{1}$}{} \hspace{0.1cm}
  Shukai~Gong\texorpdfstring{$^{1}$}{} \hspace{0.1cm}
  Daquan~Zhou\texorpdfstring{$^{1,*}$}{}
}

\affiliation[1]{Peking University}
\affiliation[2]{National University of Singapore}
\contribution[*]{%
  {\fontfamily{sansmedium}\selectfont\,Corresponding author}%
}

\checkdata[
\raisebox{-0.2em}{\includegraphics[width=0.025\linewidth]{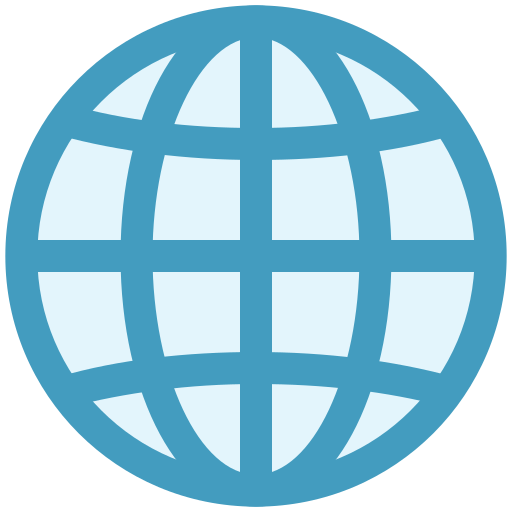}}~~Project Page]{\href{https://dagroup-pku.github.io/PyRUA-Lean/RSIJudge/}{\texttt{https://dagroup-pku.github.io/PyRUA-Lean/RSIJudge}}
\par\vspace{1.5mm}}

\checkdata[
\raisebox{-0.4em}{%
  \includegraphics[width=0.025\linewidth]{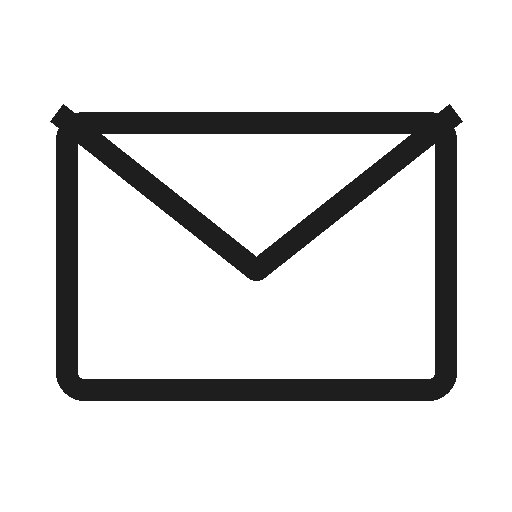}%
}~~Corresponding Author]{%
\href{mailto:zhoudaquan21@gmail.com}{\texttt{Daquan Zhou}}%
\par\vspace{1.5mm}
{%
  \fontfamily{sansmedium}\fontsize{8}{10}\selectfont
  \color{black!70}\noindent
}%
}

\abstract{
Self-improving embodied agents are increasingly being developed and compared on robot simulation benchmarks. Several
recent works expose the benchmark's success criteria to the learning agent, a practice we call Criteria Exposure (CE).
These criteria, originally written to verify policies trained by imitation learning, often demand more than the
corresponding language instructions, so a score from such a verifier may not faithfully reflect whether the instructed
task was completed, especially for robot-control programs written by an agent.
As embodied coding agents report ever higher benchmark scores, a natural question arises: when a coding agent is
exposed to the benchmark's criteria, does it generate code that better performs the instructed task, or code that
simply overfits to the verification criteria?
We conduct controlled experiments with a self-improving agent with and without CE, and find that with CE the agent
quickly produces code with a higher success rate under the given criteria, but this advantage disappears when the
criteria are changed slightly without departing from the original instructions. More importantly, the skills the agent
develops under CE transfer poorly across tasks.
We further propose instruction-faithful criteria and show that they better reflect the performance of agent-generated
programs.

}

\begin{document}
\maketitle

\suppressfloats[t]
\begin{figure}[t!]
  \centering
  \includegraphics[width=\linewidth]{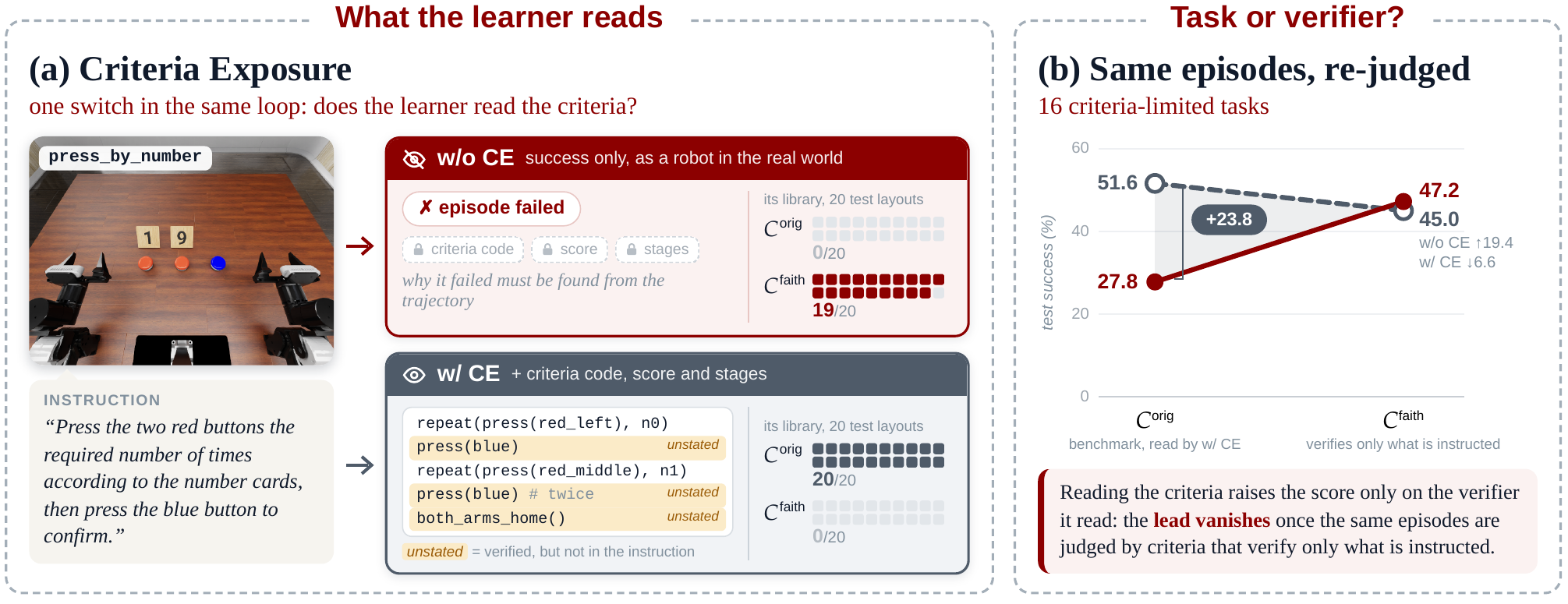}
  \caption{\textbf{Learning the task or the verifier?} \emph{(a)}: in \task{press\_by\_number}, the criteria $\Corig$
    verify a button protocol that the instruction does not state (highlighted). Without Criteria Exposure (\woCE) the
    learner sees only whether each episode succeeded; with it (\wCE) it also reads the criteria code, score and
    stages. Right of each setting: the test successes of its library on 20 layouts, judged by $\Corig$ and by the
    instruction-faithful criteria $\Cfaith$. \emph{(b)}: libraries learned under $\Corig$ on the 16 criteria-limited
    tasks, 320 test episodes each, with every episode judged by both criteria (\cref{sec:exp-rejudge}). The lead of
    \wCE under $\Corig$, \V{RJ16_ON_ORIG}\% against \V{RJ16_OFF_ORIG}\%, is gone under $\Cfaith$
    (\V{RJ16_ON_FAITH}\% against \V{RJ16_OFF_FAITH}\%).}
  \label{fig:teaser}
\end{figure}

\section{Introduction}
\label{sec:intro}

Agents that write code to control robots with multimodal language models complete many manipulation tasks
zero-shot~\citep{liang2023code,huang2023voxposer,fu2026capx,pyrualean2026}, but preserve nothing from one
episode to the next, which makes self-improvement across episodes a natural next
step~\citep{shinn2023reflexion,wang2023voyager,zhao2024expel}. Such agents are developed and compared on embodied
simulation benchmarks~\citep{chen2026robodojo,liu2023libero,nasiriany2026robocasa365} originally built to evaluate
end-to-end policies trained by imitating demonstrations, such as vision-language-action (VLA)
models~\citep{brohan2023rt2,kim2024openvla,black2025pi0,intelligence2025pi05} and world action models
(WAMs)~\citep{shen2026wamsurvey}. The actual success criteria of these benchmarks often ask for more than the
instruction; in the \task{press\_by\_number} task of RoboDojo, for instance, the criteria require the blue button to be
pressed exactly twice, between two groups of red presses, which the instruction never states (\cref{fig:teaser}a). For
policies trained by imitation, this gap is harmless, since their demonstrations already contain every detail the
criteria verify: order, counts, end states, and the return of both arms. An agent that receives only the instruction,
however, lacks this information. Existing self-improving methods therefore expose the criteria to the agent to different degrees, from rule
descriptions~\citep{chen2026eta} and published scoring tables~\citep{li2026embodiedevo,gao2026hexashell} to numeric
thresholds~\citep{huang2026rax}, formal goal predicates~\citep{wang2026harnesspai}, and the source code of the
verifier~\citep{bai2026compile}. A robot in the real world, by contrast, receives only its instruction and, at most, a success judged by a
human or a VLM. Our research question is: \emph{when a self-improving embodied agent is exposed to the
success criteria, does the code it iteratively produces better perform the instructed task, or simply overfit to the
verification criteria?}

Studies of benchmark validity and of specification gaming show that criteria can demand more than the instruction states and
that agents exploit a verifier they can
inspect~\citep{kapoor2024agents,zhu2025establishing,zhong2025impossiblebench}, but we are
not aware of work that measures how much of a reported gain in embodied self-improvement comes from reading the
criteria, or whether it survives when the criteria change. We therefore compare two settings of one fixed self-improvement loop, in which a frozen coding
agent acts and a learner agent turns its episodes into a library of skills and memory: with Criteria Exposure
(\wCE), the learner reads the criteria; without it (\woCE), the learner sees only whether each episode succeeded. We
ask three questions: on which tasks exposure raises the score; whether its lead survives a change of criteria, both of
the criteria that judge the episodes and of the criteria the learner reads; and whether what it adds carries over to
new tasks and to two further benchmarks, LIBERO-PRO and RoboCasa365. To change the criteria, we rewrite them for the RoboDojo tasks whose criteria ask for more
than the instruction, so that they verify only what the instruction asks. On these tasks every episode is judged by
both versions, and in a separate run the learner reads the rewritten criteria instead of the original ones.

Our experiments reveal three findings. First, under the original criteria, exposure helps only on tasks whose criteria
ask for more than the instruction; when criteria and instruction agree, learning from success alone does just as well. Second, the score of exposure follows the criteria
it read: when the same episodes are judged by instruction-faithful criteria, its advantage disappears, as the exposed
libraries lose successes while the success-only libraries gain them (\cref{fig:teaser}b); conversely, a library learned from
the rewritten criteria scores far lower when judged by the original ones. Third, the lead of exposure does not reappear on new RoboDojo tasks, and across benchmarks neither the exposed nor the
success-only library gives a clear gain over the empty library; only experience gathered within the target benchmark
does.

Taken together, these results suggest a practical principle for evaluating embodied self-improvement: the learner
should receive only the feedback a robot would receive in the real world, and the benchmark's criteria should verify what the
instruction says. A gain that holds when the criteria are changed but still agree with the instruction reflects what
the agent has learned about the task. Seen
this way, whether the learner reads the criteria is not an implementation detail of a self-improving agent but a
choice about measurement, and it determines what a reported gain means.

\section{Related Work}
\label{sec:related}

\claim{Code-writing embodied agents and their benchmarks.}Language models plan and act for robots by selecting skills~\citep{ahn2022saycan}, writing policy
code~\citep{liang2023code}, or composing value maps~\citep{huang2023voxposer}; recent agents close the loop with code
replanning and thin tool layers~\citep{fu2026capx,pyrualean2026,gao2026hexashell}, and frontier models now control
robots zero-shot. These agents are evaluated on benchmarks built for vision-language-action
policies trained on demonstrations~\citep{brohan2023rt2,kim2024openvla,black2025pi0,intelligence2025pi05}:
RoboDojo~\citep{chen2026robodojo},
LIBERO and LIBERO-PRO~\citep{liu2023libero,zhou2025liberopro}, RoboCasa365~\citep{nasiriany2026robocasa365}, and
RoboTwin~2.0~\citep{chen2025robotwin2}. These works score agents with criteria written for demonstration-trained
policies; whether the criteria verify only what the instruction asks is rarely examined on its own, and we measure how this gap
enters the score of a self-improving agent.

\claim{Self-improving agents and the feedback their learners see.}In digital domains, programs are evolved against automatic
evaluators~\citep{novikov2025alphaevolve,zhang2025dgm}, agents reflect on textual
feedback~\citep{shinn2023reflexion,zhao2024expel}, skills are verified in a game
world~\citep{wang2023voyager}, and code is revised against tests~\citep{yang2024sweagent}. Embodied systems differ in
how much of the success criteria their learners or agents see. Some give only success or keep
the grader isolated~\citep{shen2026embodiedswe,lu2026aspire}; others add rule descriptions with coarse
feedback~\citep{chen2026eta}, published scoring tables and partial numeric
thresholds~\citep{li2026embodiedevo,gao2026hexashell,huang2026rax}, formal goal predicates~\citep{wang2026harnesspai},
or the source code of the verifier~\citep{bai2026compile}. In RATs, the verifier reads goal
predicates and ground truth while the policy writer's side is filtered~\citep{zhang2026rats}. In these systems the
amount of criteria information is mostly fixed by design rather than varied as an experimental factor. The closest comparison is an
ablation of EmbodiedEvo between task descriptions alone and descriptions with the wiki, but both arms are judged by the
original criteria~\citep{li2026embodiedevo}. We treat exposure as the variable and test what it gains under a second set
of criteria.

\claim{Specification gaming and benchmark validity.}Optimizing a proxy diverges from the intended objective~\citep{amodei2016concrete,skalse2022reward}, with measurable
effects of reward misspecification~\citep{pan2022effects}. Language agents exploit tests and evaluators they can
inspect~\citep{zhong2025impossiblebench,bondarenko2025demonstrating}, and agentic benchmarks need checks of their own
validity~\citep{kapoor2024agents,zhu2025establishing}. Where rule-based criteria are missing, success is judged by
VLMs~\citep{du2023success} or by autonomous real-world evaluation~\citep{zhou2025autoeval}. Building
on these studies, we further show that on embodied benchmarks the act of reading the criteria turns the gap between
criteria and instruction into score, and we measure this share with counterfactual criteria that score the same
instructions.

\section{Preliminaries and Study Design}
\label{sec:design}

This section describes the controlled study we use to analyze what a self-improving embodied agent learns from the
success criteria. The study compares two settings of one loop, \wCE and \woCE, and answers three questions: where
exposure raises the score under the original criteria, whether the advantage holds when the criteria change, and
whether the library carries over to new tasks and benchmarks. We describe in turn the loop and its two settings,
criteria-limited tasks, instruction-faithful criteria, and the evaluation protocol.

\subsection{Self-Improvement Loop and Criteria Exposure}
\label{sec:loop}

\begin{figure}[t]
  \centering
  \includegraphics[width=\linewidth]{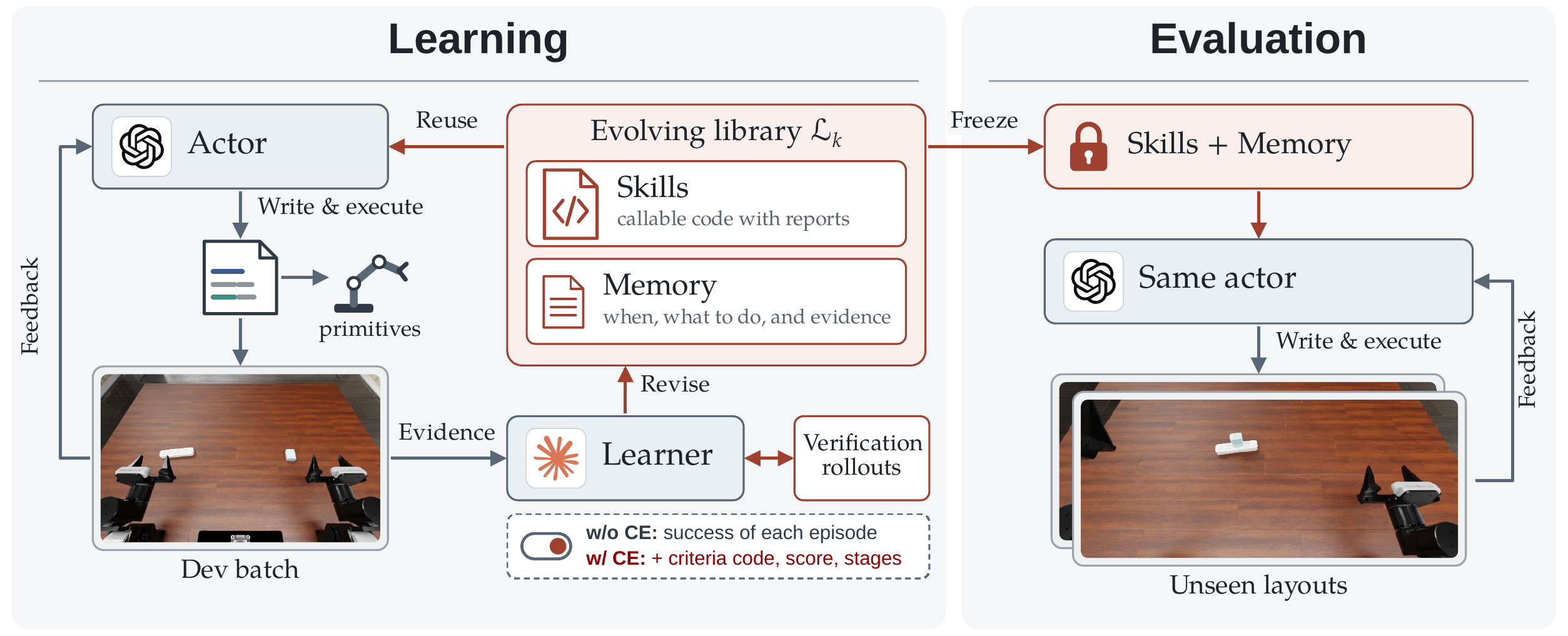}
  \caption{\textbf{Overview of the self-improvement loop.} \emph{Left, learning:} the actor (GPT-6 Astra, high reasoning
    effort) writes and executes Python cells that call the robot's primitives on five development layouts, and sees the
    result of each cell. Between batches the learner (Claude Code with Opus~5.5, high reasoning effort) turns the evidence into
    one revision of the library $\mathcal{L}_k$, whose skills are callable code that reports what it did and whose
    memory entries say when to apply what, with the evidence for it; a revision is usually tried first in verification
    rollouts, and the loop rolls back to the best revision so far when a batch falls clearly below it. Whether the
    learner also sees the success criteria is the one switch between the two settings, \wCE and \woCE (the switch
    under the learner).
    \emph{Right, evaluation:} the final library $\mathcal{L}_K$ is frozen and read-only, and the same actor runs alone
    on layouts the loop never saw.}
  \label{fig:design}
\end{figure}

A task $\tau=(I_\tau, D_\tau, C_\tau)$ has an instruction $I_\tau$, a distribution of layouts $D_\tau$ and success
criteria $C_\tau$ that judge each trajectory; some benchmarks also report a score and stages. The success the
instruction asks for, $S_\tau$, is defined by $I_\tau$ and cannot be computed directly. A frozen actor $\pi$, GPT-6
Astra with high reasoning effort, acts through the primitives of PyRUA-Lean~\citep{pyrualean2026}.
In each round a learner $U$, Claude Code with Opus~5.5 and high reasoning effort, makes one revision of the library $\mathcal{L}$, a set of
callable skills and memory entries, from the evidence $E_v(B)$ of the last batch, and the revision is rolled back if the
batch clearly regresses. Each task has its own library, learned over five rounds on five development layouts.
Algorithm~\ref{alg:loop} and \cref{fig:design} summarize the loop for one task.

\label{sec:settings}The loop runs in two settings. Without Criteria Exposure (\woCE), the learner sees only whether
each episode succeeded; the criteria code, the score and the stages are withheld. A robot in
the real world has no criteria code and receives success from a human or a VLM, so this is the setting in simulation
closest to the real world. With Criteria Exposure (\wCE), the learner also reads the criteria code of the task and the
score and stages of each episode, as systems that give the learner the source of the verifier do; the actor never reads them. The actor, the primitives, perception and benchmark scoring are frozen
and only the library changes; the two settings share the loop and differ in what the learner reads, so
differences between them reflect exposure rather than the loop itself.

\begin{algorithm}[!hb]
  \caption{Self-improvement loop}
  \label{alg:loop}
  \small
  \begin{algorithmic}[1]
    \Require actor $\pi$; dev layouts $D_{\text{dev}}$; rounds $K$; setting $v\in\{\text{\wCE},\text{\woCE}\}$; margin $\theta=2$
    \State $\mathcal{L}_0 \gets (S_0, M_0) = (\varnothing, \varnothing)$ \Comment{skills and memory}
    \State $\mathcal{L}^\star \gets \mathcal{L}_0$;\; $H \gets [\,]$;\; $B_0 \gets \textsc{Run}(\pi, \mathcal{L}_0, D_{\text{dev}})$ \Comment{zero-shot batch}
    \For{$k = 1, \dots, K$}
      \State $E \gets \textsc{Evidence}(B_{k-1}, v)$ \Comment{trajectories; criteria, score and stages only under \wCE}
      \State $(\Delta_S, \Delta_M, h) \gets \textsc{Improve}(E, H, \mathcal{L}_{k-1})$ \Comment{edit skills, memory or both; verify by rollouts}
      \State $\mathcal{L}_k \gets \textsc{Freeze}(S_{k-1} \oplus \Delta_S,\, M_{k-1} \oplus \Delta_M)$ \Comment{structural checks only, not success}
      \State $B_k \gets \textsc{Run}(\pi, \mathcal{L}_k, D_{\text{dev}})$
      \If{$B_k$ has $\geq \theta$ fewer successes than the batch of $\mathcal{L}^\star$} \State $\mathcal{L}_k \gets \mathcal{L}^\star$ \Comment{rollback}
      \ElsIf{$B_k$ is the best batch so far} \State $\mathcal{L}^\star \gets \mathcal{L}_k$
      \EndIf
      \State append $(\Delta_S, \Delta_M, B_k, h)$ to $H$ \Comment{change, result, hypothesis}
    \EndFor
    \State \Return $\mathcal{L}_K$ \Comment{the active library after round $K$}
  \end{algorithmic}
\end{algorithm}

\subsection{Criteria-limited Tasks}
\label{sec:limited}

We call a task \emph{criteria-limited} if $C_\tau$ disagrees with $S_\tau$ on some trajectories, either by rejecting a
trajectory $S_\tau$ accepts or by accepting one $S_\tau$ rejects (criteria looser than the instruction); all other
tasks are \emph{faithful}. Such tasks appear in every benchmark we examined, RoboDojo, LIBERO~\citep{liu2023libero}
(whose instructions and goals LIBERO-PRO keeps), RoboCasa365 and RoboTwin~2.0~\citep{chen2025robotwin2}, and their
disagreements fall into a few kinds. The criteria may ask for more than the instruction states, such as an order, a
count or an end state: in RoboDojo, \task{press\_by\_number} requires a button protocol its instruction never gives.
They may be stricter than the instruction needs, such as an unstated tolerance for an object resting on another in
LIBERO. They may contradict the instruction, as in a LIBERO task that asks to stack a bowl on the front bowl while its
criteria require the back one. Or they may verify less than the instruction asks: in RoboTwin~2.0,
\task{shake\_bottle\_horizontally} succeeds once the bottle is lifted.

We rewrite criteria and re-judge episodes on RoboDojo only. Its criteria report a score and stages
besides success, so an exposed learner reads the most complete information there, and we compared the criteria of all
of its tasks with their instructions; rewriting the criteria of other benchmarks and repeating these experiments on
them is left to future work. We distinguish seven types: hidden
rules, hard thresholds, timing and irreversible events, criteria defects, extra end states, loose criteria,
and contradictions. In RoboDojo, after a manual review of every task, the \V{AUDIT_CL_TASKS} criteria-limited tasks have
\V{AUDIT_CL_VARIANTS} variants, and hidden rules and hard thresholds are the most common types. Criteria-limited is defined with respect to the
instruction the agent receives at reset; the RoboDojo website publishes part of these rules in its task descriptions and
scoring tables.

\subsection{Instruction-faithful Criteria}
\label{sec:faithful}

For each criteria-limited task we rewrite only the criteria code, never the instruction, and leave the original
repository untouched: hidden rules are removed or relaxed, defects fixed, hard thresholds relaxed to what still clearly
satisfies the instruction, loose criteria tightened where the simulator state can detect the difference, and the
return of both arms kept. The two criteria judged the same episodes independently, and every episode
whose verdict flipped was checked by hand without finding a misjudgment by the new criteria.
Each evaluation episode is then judged by both criteria, giving $\SRo$ and $\SRf$. Since both criteria score the same
instruction on the same episode, any difference between the two success rates comes from what the criteria verify, not
from behavior. $\Cfaith$ is ours and serves as a proxy for $S_\tau$; we discuss this in
\cref{sec:limitations}.

\subsection{Experiment Settings and Evaluation Protocol}
\label{sec:protocol}

To isolate the effect of exposure, the two settings of each comparison use the same actor, primitives, round budget
and development layouts. The \V{LEARN_N} learned tasks comprise \V{LEARN_CL} criteria-limited and \V{LEARN_FAITH}
faithful tasks, grouped in advance by whether their criteria were rewritten; Appendix~\ref{app:tasks} lists the tasks. Development uses layouts 0--4 of seed~1 and
testing layouts 0--19 of seed~0. We report the success rate (SR) and compare
settings layout by layout. All experiments use the same actor and learner.
\label{sec:reuse}To reuse libraries, we distill the libraries of ten source tasks into one shared library and apply it to
disjoint target tasks, in RoboDojo and on two further benchmarks with aligned primitives.

We use three simulated benchmarks with different roles. \textbf{RoboDojo} is the main benchmark: its rule-based criteria
report a score and stages, and all three questions are answered on it. \textbf{LIBERO-PRO} is a single-arm tabletop
benchmark where zero-shot success is already near the ceiling; it tests whether transfer still has room at high scores.
\textbf{RoboCasa365} has household kitchen scenes and the lowest zero-shot scores; it tests cross-benchmark experience
and also hosts in-benchmark learning.

\section{Experiments and Analysis}
\label{sec:experiments}

We test three hypotheses:
\begin{itemize}[leftmargin=*, itemsep=2pt]
  \item \textbf{H1.} Exposure raises the score only on tasks whose criteria ask for more than the instruction
  (\cref{sec:exp-orig}).
  \item \textbf{H2.} If exposure teaches the verifier rather than the task, its lead shrinks or reverses when the same
  episodes are judged by instruction-faithful criteria (\cref{sec:exp-rejudge}), and each exposed library scores best
  under whichever criteria its learner read (\cref{sec:exp-swap}).
  \item \textbf{H3.} What exposure adds does not carry over to new tasks or benchmarks (\cref{sec:exp-reuse}).
\end{itemize}

\subsection{Exposure under the Original Criteria}
\label{sec:exp-orig}

We first judge the benchmark's way, by the original criteria, which are also the criteria the exposed learner reads.

\subsubsection{Success-only versus Exposed Self-Improvement}
\label{sec:exp-orig-main}

\paragraph{Setup.} We learn one library per task and setting on the \V{LEARN_N} learned tasks (\V{LEARN_CL}
criteria-limited, \V{LEARN_FAITH} faithful), over five rounds on five development layouts, and test each library on
layouts 0--19 of seed~0 against the empty library. The two groups were fixed before any result was seen.

\begin{table}[!hb]
  \centering
  \caption{\textbf{Test success under the original criteria} (SR, \%, layouts 0--19 of seed~0). Parentheses: difference
    from the empty library in points. The criteria-limited tasks count the episodes of \cref{sec:exp-rejudge}, judged
    by $\Corig$.}
  \label{tab:orig}
  \small
  \begin{tabular}{@{}l c c c@{}}
    \toprule
    Tasks & Empty library $\Lempty$ & \woCE & \wCE \\
    \midrule
    All (\V{LEARN_N}) & \V{T_ALL_V0} & \V{T_ALL_OFF} (\V{T_ALL_OFF_D}) & \V{T_ALL_ON} (\V{T_ALL_ON_D}) \\
    Criteria-limited (\V{LEARN_CL}) & \V{T_CL_V0} & \V{T_CL_OFF} (\V{T_CL_OFF_D}) & \V{T_CL_ON} (\V{T_CL_ON_D}) \\
    Faithful (\V{LEARN_FAITH}) & \V{T_FA_V0} & \V{T_FA_OFF} (\V{T_FA_OFF_D}) & \V{T_FA_ON} (\V{T_FA_ON_D}) \\
    \bottomrule
  \end{tabular}
\end{table}

\begin{figure}[t]
  \centering
  \begin{minipage}[t]{0.53\linewidth}
    \vspace{0pt}
    \includegraphics[width=\linewidth]{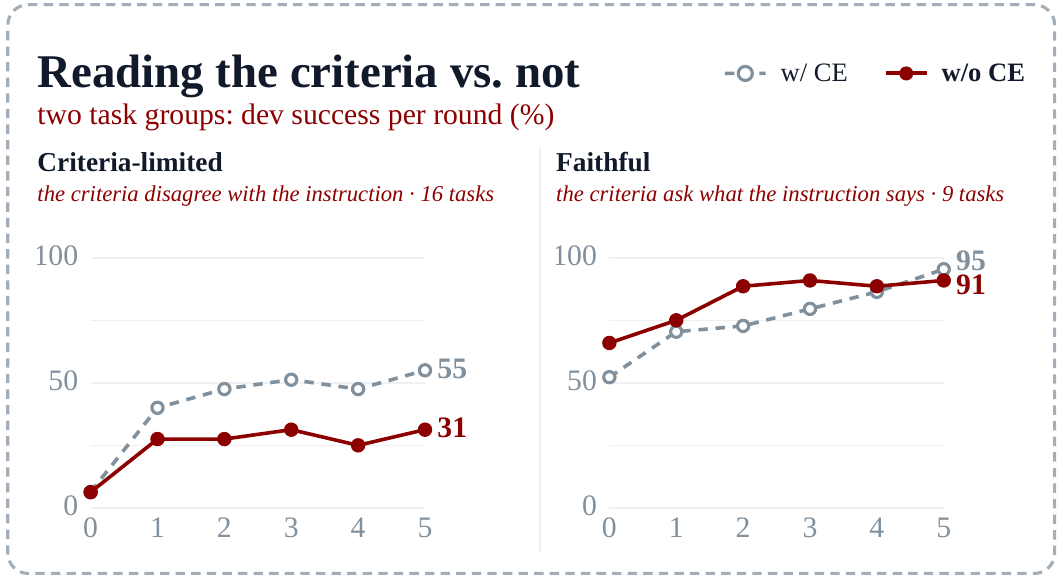}
  \end{minipage}\hfill
  \begin{minipage}[t]{0.44\linewidth}
    \vspace{0pt}
    \captionsetup{skip=0pt}
    \caption{\textbf{Development success by round}, on the \V{LEARN_CL} criteria-limited and the \V{LEARN_FAITH}
      faithful tasks. Dashed: \wCE; solid: \woCE; round 0 is the empty library. On criteria-limited tasks both
      settings start at 6\% and part in round 1 (40\% against 28\%); after that the success-only learner barely
      moves and ends at 31\%, against 55\% with exposure. On faithful tasks the success-only learner is ahead
      through round 4 and both end above 90\% (91\% and 95\%); its lead, at most 16 points, is close to the
      14-point gap of round 0 (66\% against 52\%), where both run the same empty library.}
    \label{fig:dev-curves}
  \end{minipage}
\end{figure}

\paragraph{Results.} \Cref{tab:orig} and \cref{fig:dev-curves} show three patterns. First, learning from success
alone doubles success, from \V{T_ALL_V0}\% to \V{T_ALL_OFF}\%, and reading the criteria adds another \V{ON_GAIN}~points
(\V{T_ALL_ON}\%). Second, this extra gain comes from the criteria-limited tasks: on faithful tasks the two settings are
level (\V{T_FA_OFF}\% and \V{T_FA_ON}\%), while on criteria-limited tasks exposure leads, \V{T_CL_ON}\% against
\V{T_CL_OFF}\%. Third, among criteria-limited tasks the lead depends on whether an unstated rule shows in failed
episodes: in \task{build\_tower} the leftover blocks reveal it and both settings reach 20/20, whereas the button protocol of
\task{press\_by\_number}, blue pressed twice and between the two red groups, never shows, and the success-only
learner never finds it.

\subsubsection{Synthesis}
\label{sec:exp-orig-syn}

\begin{itemize}[leftmargin=*, itemsep=4pt]
  \item \textbf{Exposure pays off only where the criteria ask for more than the instruction.} Richer feedback is often
  assumed to make self-improvement better. Our results show a condition: when the criteria verify what the instruction
  says, success alone tells the learner what the task requires, and what remains, a reliable procedure, is learned
  equally well in both settings. Exposure helps only with rules that the instruction omits and that failures do not
  reveal.

  \item \textbf{The advantage of exposure measures the benchmark as much as the agent.} Because it comes from what the
  instruction leaves out, the gain of exposure grows with the gap between a benchmark's instructions and its criteria.
  Some of what the criteria reveal is legitimate task knowledge, but it belongs in the instruction: benchmarks that
  state in the instruction what their criteria verify (counts, order, end states, the reset of the arms) and release
  versioned criteria with a comparison against the instructions would let self-improvement be compared on
  the task alone.
\end{itemize}

\subsection{Changing the Criteria: Task or Verifier?}
\label{sec:exp-change}

\Cref{sec:exp-orig} judged with the criteria the exposed learner read. Changing them separates two explanations. If
exposure teaches the task, its lead should survive under $\Cfaith$; if it teaches the verifier, the lead should shrink or
reverse, and each library should score best under whichever criteria its learner read. We change the criteria used
for judging and the criteria that are read.

\subsubsection{Re-judging with Instruction-faithful Criteria}
\label{sec:exp-rejudge}

\paragraph{Setup.} We run the libraries of \cref{sec:exp-orig-main} on the 16 learned criteria-limited tasks, all of
which have rewritten criteria, 20 layouts each, and judge every episode by both criteria.

\begin{figure}[H]
  \centering
  \begin{minipage}[t]{0.49\linewidth}
    \centering
    \includegraphics[width=\linewidth]{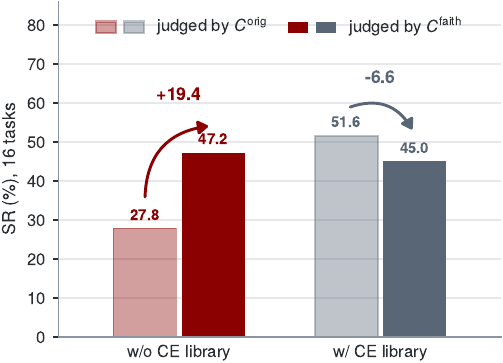}
    \caption{\textbf{The same episodes judged by the original and the instruction-faithful criteria.} Libraries
      learned under $\Corig$ on the 16 criteria-limited tasks, 320 episodes per library. Light bars: judged by $\Corig$;
      solid bars: the same episodes judged by $\Cfaith$; arrows give the change in points.}
    \label{fig:rejudge}
  \end{minipage}\hfill
  \begin{minipage}[t]{0.49\linewidth}
    \centering
    \includegraphics[width=\linewidth]{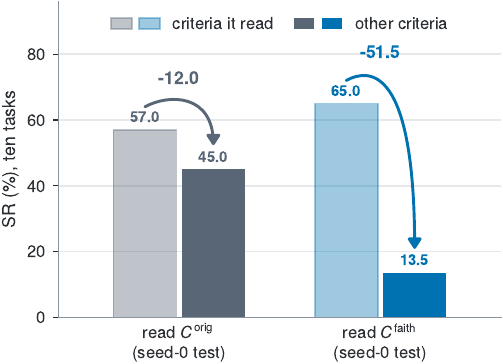}
    \caption{\textbf{Each exposed library judged by the criteria it read and by the other.} Both on the ten tasks
      of $\LCEf$, seed-0 test, 200 episodes per library. $\LCEo$: their episodes in \cref{fig:rejudge}. $\LCEf$: its
      final library on the same layouts. Light bars: judged by the criteria the learner read; solid bars: the same episodes judged by the other
      criteria; arrows give the change in points.}
    \label{fig:matrix}
  \end{minipage}
\end{figure}

\paragraph{Results.} \Cref{fig:rejudge} shows the lead of exposure disappear. Under $\Corig$ the exposed library
leads, \V{RJ16_ON_ORIG}\% against \V{RJ16_OFF_ORIG}\%: paired by task and layout, it alone succeeds on
\V{RJ16_ORIG_ONONLY} layouts and the success-only library alone on \V{RJ16_ORIG_OFFONLY}. Judged by $\Cfaith$, the same episodes give \V{RJ16_ON_FAITH}\% for the exposed library and
\V{RJ16_OFF_FAITH}\% for the success-only one, a difference within noise (\V{RJ16_FAITH_ONONLY} against
\V{RJ16_FAITH_OFFONLY} layouts); the gap shrinks by \V{RJ16_GAPCHG}~points (95\% bootstrap interval over
tasks \V{RJ16_GAPCHG_LO}--\V{RJ16_GAPCHG_HI}). The success-only library loses almost nothing:
$\Cfaith$ accepts all but 2 of its 89 successes under $\Corig$, and it gains where the original criteria demand an
unstated protocol, as in \task{press\_by\_number}, \task{swap\_blocks} and \task{make\_kong}. The exposed library loses
where it had learned that protocol, and on tasks whose criteria changes mostly fix defects or relax thresholds the two
criteria give nearly the same scores.

\subsubsection{Swapping the Criteria that Are Read}
\label{sec:exp-swap}

\paragraph{Setup.} We then let the learner read $\Cfaith$ instead (ten tasks, five rounds)
and compare the resulting library $\LCEf$ with $\LCEo$ on the same ten tasks, each judged by the criteria its learner
read and by the other.

\paragraph{Results.} \Cref{fig:matrix} shows that each exposed library does well only under the criteria it read:
$\LCEo$ falls from \V{SW_ON_ORIG}\% to \V{SW_ON_FAITH}\% under $\Cfaith$, and $\LCEf$ from \V{SW_TEST_FAITH}\% to
\V{SW_TEST_ORIG}\% under $\Corig$. The fall is steeper for $\LCEf$ because the
two directions differ: $\Cfaith$ mostly removes or relaxes the hidden rules and thresholds of $\Corig$, so moving $\LCEo$
to $\Cfaith$ loosens the verification, whereas moving $\LCEf$ to $\Corig$ adds back every rule the library never saw. The score
follows the criteria that were read even where they are looser: having read $\Cfaith$, the library no longer sets the
bottle upright in \task{pour\_by\_language}, which $\Corig$ requires, and solves 19 of its 20 test layouts under
$\Cfaith$ but none under $\Corig$. This does not tell which criteria are right, which
\cref{sec:exp-rejudge} addresses, but it shows that the score of exposure does not carry between two criteria that
score the same instruction.

\subsubsection{Synthesis and Implications}
\label{sec:exp-change-syn}

\begin{itemize}[leftmargin=*, itemsep=4pt]
  \item \textbf{Exposure makes self-improvement fit the criteria rather than learn the instructed task.} A higher
  score from reading the criteria is commonly taken as faster learning of the task. A learner that reads the
  criteria, however, revises the library until it passes the verifier it reads, so the library learns the task only as
  far as the criteria agree with the instruction. Under instruction-faithful criteria the exposed library loses
  successes that the success-only library gains, and each exposed library scores best only under the criteria it read:
  on \task{press\_by\_number} it solves all 20 layouts under $\Corig$ and none under $\Cfaith$, and the library that read
  $\Cfaith$ solves \V{SW_TEST_FAITH}\% of its test episodes under $\Cfaith$ but \V{SW_TEST_ORIG}\% under $\Corig$.
  Self-improvement should therefore be measured without exposure by default, with the success rate under
  instruction-faithful criteria reported alongside: a robot in the real world has no criteria code, so the part of a
  gain that survives a change of criteria is the part to expect there.

  \item \textbf{Exposure changes what the learner writes down.} A learner that reads the criteria can explain a
  failure by the condition it violated, so its experience is phrased in the terms of one verifier; a learner that sees
  only success has to find a physical cause, and its rules are closer to regularities of the robot. About a third of
  the exposed memory entries cite the criteria code, and none of the success-only ones do.
\end{itemize}

\subsection{Reusing the Library: What Carries Over}
\label{sec:exp-reuse}

Finally, we carry the libraries to new tasks in RoboDojo and to two further benchmarks.

\subsubsection{Across Tasks within RoboDojo}
\label{sec:exp-tasks}

\paragraph{Setup.} We distill the libraries of ten learned tasks into one library per setting and
apply it to ten target tasks that none of the source tasks covers,
paired with the empty library on the same layouts.

\begin{table}[!hb]
  \centering
  \caption{\textbf{Reusing libraries on new tasks and benchmarks} (successes, each benchmark's own criteria). The
    RoboDojo libraries are distilled from ten learned tasks and carried over to LIBERO-PRO and RoboCasa365; the
    last row uses libraries distilled from other RoboCasa365 tasks. Every row is paired with the empty library on the
    same layouts; parentheses give the change in successes against it.}
  \label{tab:transfer}
  \small
  \begin{tabular}{@{}l l c c c@{}}
    \toprule
    Target & Library source & $\Lempty$ & \woCE & \wCE \\
    \midrule
    RoboDojo, 10 target tasks $\times$ 20 layouts & RoboDojo & 65/200 & 80/200 \dpos{15} & 71/200 \dpos{6} \\
    LIBERO-PRO, 6 tasks $\times$ 10 seeds & RoboDojo & 45/60 & 48/60 \dpos{3} & 46/60 \dpos{1} \\
    RoboCasa365, 10 target tasks $\times$ 10 seeds & RoboDojo & 22/100 & 24/100 \dpos{2} & 24/100 \dpos{2} \\
    RoboCasa365, 10 target tasks $\times$ 10 seeds & RoboCasa365 & 22/100 & 33/100 \dpos{11} & 29/100 \dpos{7} \\
    \bottomrule
  \end{tabular}
\end{table}

\paragraph{Results.} Both settings help on the target tasks (first row of \cref{tab:transfer}): the empty library
solves 65 of 200 episodes, the success-only library 80 and the exposed one 71. The \V{ON_GAIN}-point lead of exposure on
its own tasks does not reappear.

\subsubsection{Across Benchmarks}
\label{sec:exp-bench}

\paragraph{Setup.} We apply the same libraries, translated to each target's prompt and primitives, on LIBERO-PRO and
RoboCasa365, whose primitives we aligned with RoboDojo's. On LIBERO-PRO the empty library
already solves \V{LB_FULL} of the \V{LB_TASKS} tasks on all five seeds, so we use the \V{LB_NOTFULL} others, with ten
seeds each. On RoboCasa365 we also distill one library per setting from other RoboCasa365 tasks, as an in-benchmark
reference.

\paragraph{Results.} The libraries from RoboDojo add little on either benchmark (\cref{tab:transfer}): 48 and 46 of 60
episodes against 45 on LIBERO-PRO, and 24 and 24 of 100 against 22 on RoboCasa365. The libraries distilled within
RoboCasa365 do better, 33 (\woCE) and 29 (\wCE) of 100.

\subsubsection{Synthesis and Implications}
\label{sec:exp-reuse-syn}

\begin{itemize}[leftmargin=*, itemsep=4pt]
  \item \textbf{What exposure adds is specific to the task it was learned on.} One might expect the setting that scores
  higher on its own tasks to transfer better. It does not: the lead of exposure disappears on target tasks. Exposed experience carries the conditions and precision needed
  to pass one task's criteria, which become noise or overhead elsewhere, whereas rules about the gripper hold across
  tasks.

  \item \textbf{Experience transfers where the bottleneck is shared.} Within a benchmark the robot, the scenes and
  perception stay the same, and learned rules carry over. Across benchmarks the general part of the experience is
  already covered by the target's prompt, while what the target lacks has no counterpart in the source, so the
  transferred skills go unused; experience gathered on the target benchmark, even on other tasks, addresses its own
  bottleneck and helps.
\end{itemize}

\section{Limitations and Future Work}
\label{sec:limitations}

\claim{The instruction-faithful criteria are ours, and the feedback is simulated.} $\Cfaith$ serves as our proxy for
the success the instruction asks for, $S_\tau$. Neither setting's libraries saw the rewritten criteria when they were
written; rules and types were fixed in advance; a second seed was judged independently and every flipped episode
checked. Still, only RoboDojo's criteria were rewritten,
and success-only feedback in simulation is cleaner than the success a human or a VLM reports for a robot in the real
world, which has no criteria code. Next steps are an independent review, released criteria code, rewritten criteria for a second
benchmark, and VLM success detection in place of the simulator's success~\citep{du2023success}.

\claim{Exposure is studied at two levels, as current systems implement it.} Most existing systems sit at intermediate
levels (rule descriptions, published scoring tables, partial thresholds), which we did not study separately. A next step
is to expose these levels one at a time and measure how the gain under $\Corig$, and its loss under $\Cfaith$, grow
with what the learner reads.

\claim{Samples are small and the mechanisms are inferred.} Each task has 20 test layouts, so quantitative claims use
pooled results and single tasks are described only qualitatively; and distillation variance was not estimated. The accounts of what exposure writes into the library, of what carries over and of how the actor uses transferred libraries rest on entry text, skill calls and
action statistics, not on ablations. Next steps are repeated distillations and ablations that remove or swap individual
entries, such as the re-measuring entries of the distilled \wCE library or the memory entries of a
translated library.

\section{Conclusion}
\label{sec:conclusion}

In this paper, we studied what a self-improving embodied agent learns when it reads the success criteria, by comparing
one self-improvement loop with and without Criteria Exposure, under the benchmark's own and instruction-faithful
criteria, and on new tasks and benchmarks. Our results show that self-improvement from success alone is real and
survives a change of criteria, whereas the score that Criteria Exposure adds comes mainly from satisfying the criteria
that were read: it appears only where the criteria and the instruction disagree, disappears under instruction-faithful
criteria, and across benchmarks neither the exposed nor the success-only library gives a clear gain, while only
in-benchmark experience helps. These findings
provide a more careful reading of reported self-improvement gains on embodied benchmarks and practical guidance for
measuring them: use the feedback a robot receives in the real world, and judge by criteria that verify what the instruction says.
A robot in the real world never sees its verifier; the agent we measure should not either.

\clearpage
\bibliographystyle{plainnat}
\setlength{\bibhang}{0pt}
\setlength\bibindent{0pt}
\bibliography{references}

\clearpage
\appendix

{
  \centering
  \textbf{\Large Learning the Task or Learning the Verifier?\\[2pt] Rethinking Criteria Exposure in Embodied Self-Improvement}

  \vspace{5pt}
  {\Large Appendix}
  \par
}

\renewcommand{\topfraction}{0.9}\renewcommand{\textfraction}{0.07}\renewcommand{\floatpagefraction}{0.85}
\section{Implementation Details}
\label{app:implementation}

\subsection{Experimental protocol}
\label{app:protocol}

\claim{Benchmark.} RoboDojo's simulation track uses a dual-arm ARX-X5 with RGB observations only; its 42 tasks span
five capability dimensions. The \V{TEST_TASKS_N} tasks on which the zero-shot actor solves fewer than half of the
episodes of the official evaluation seed, with five stronger ones, make the \V{LEARN_N} compared tasks
(\cref{tab:tasks}). Every run
uses one version of RoboDojo (Table~\ref{tab:settings}); task source, criteria, leaderboard script and evaluation
layouts are identical file by file across the checkouts we compare.

\claim{Conditions.} The actor's model, primitives and prompt are the same in every condition. The learner with
Criteria Exposure reads the criteria code, each episode's score and its passed stages; the one without sees full
trajectories and success only. All conditions are compared over five rounds of improvement (rounds 0--5). The groups
(Appendix~\ref{app:audit}) rest on reading the code, the learners'
diagnoses and the failed rollouts.

\claim{Layouts.} Development reruns layouts 0--4 of official evaluation seed 1 every round; the test layouts are
layouts 0--19 of seed 0 (\V{HO_ALL_EPS} episodes per condition), which no learner sees.

\claim{Statistics.} Development batches are noisy: rerunning one library on one layout changes the verdict in
\V{RERUN_DIFFER_PCT}\% of cases, and the two round-0 batches of the same empty library differ in at least
\V{R0_CROSS_DIFFER_PCT}\% of episodes. We therefore draw conclusions from the test layouts and pooled curves, and read
single-task development curves only qualitatively.

\subsection{Implementation and settings}
\label{app:impl}

\textbf{Actor.} GPT-6 Astra writes one code cell per MLLM call against PyRUA-Lean's frozen \texttt{robo}
API~\citep{pyrualean2026}: RGB perception (SAM~3~\citep{carion2025sam3} and multi-view measurement) and motion
primitives, with no VLA policy, depth or simulator state. The library is installed as a package \texttt{library/}:
cells import its skills, and its live memory entries are appended to the prompt.

\textbf{Library.} A skill is a function of \texttt{robo} that imports only standard modules, \texttt{numpy} and other
library modules, does no I/O and returns \texttt{\{"ok": ..., "why": ..., ...\}}. A memory entry is
\texttt{\#\# \{task\}-<stage>-<nnn>} followed by stage, symptom, applies, do, status and evidence. A task's library
holds at most 40,000 characters, all of which the actor reads.

\textbf{Learner.} Claude Code reads the session, the task's dev runs, the API source and, with Criteria Exposure,
the task's criteria; it writes only the candidate library, its verification rollouts and its notes. Without Criteria
Exposure, criteria, scores and stages are removed from the evidence.

\textbf{Acceptance and rollback.} A candidate may change only the task's own directory, must keep every memory entry
identifier, must cite a judged rollout or dev batch for each entry newly marked \texttt{dev\_verified}, and must pass
the static skill rules and the size limit; none of this looks at success, and no candidate was refused. The active
revision rolls back to the best earlier complete batch when a complete batch has two or more fewer successes; ties go
to the batch that passed more stages, and a batch with an episode lost to infrastructure neither triggers a rollback
nor becomes the best.

\begin{table}[!htbp]
  \centering
  \caption{Settings (the same in every run).}
  \label{tab:settings}
  \footnotesize
  \begin{tabular}{@{}lp{0.66\linewidth}@{}}
    \toprule
    \multicolumn{2}{@{}l}{\emph{Benchmark}} \\
    Code & RoboDojo commit \texttt{cb2a37d58dbb} (2026-10-08), XPolicyLab \texttt{53c5da5a01bc} \\
    Assets & HF dataset \texttt{RoboDojo-Benchmark/RoboDojo}, revision \texttt{35efbc7dedfd} (2026-09-28) \\
    Simulator & Isaac Sim 5.1.0, Isaac Lab 0.54.3; suite ARX-X5, RGB only \\
    \midrule
    \multicolumn{2}{@{}l}{\emph{Actor}} \\
    Model & GPT-6 Astra, reasoning effort high, at most 8,192 output tokens per request \\
    Interface & PyRUA-Lean cells, RGB perception, no VLA, ARX-X5 suite \\
    Per episode & at most 100 cells, 200 MLLM requests, 20 nudges, 90 minutes of policy time; the task's own
      native-action limit \\
    \midrule
    \multicolumn{2}{@{}l}{\emph{Learner}} \\
    Agent & Claude Code, Claude Opus 5.5, effort high \\
    Per session & up to 2 attempts of 120 minutes; at most 8 rollouts, 3 per dev layout \\
    Library & task-local skills and memory, at most 40,000 characters per task \\
    \midrule
    \multicolumn{2}{@{}l}{\emph{Campaign}} \\
    Splits & dev: seed 1, layouts 0--4; test: seed 0, layouts 0--19 \\
    Rounds & 6 including round 0 (five rounds of improvement) \\
    Rollback & to the best complete earlier batch, if a complete batch has $\geq 2$ fewer successes \\
    \bottomrule
  \end{tabular}
\end{table}

\subsection{Evaluation details}
\label{app:eval}
The \V{LEARN_N} learned tasks are grouped by whether their criteria were rewritten. Layouts that the benchmark marks unstable are replaced by
the next layout in the official order. Under dual judging an episode ends early only when both criteria have reached a
verdict, so neither verdict is truncated.

\subsection{Tasks}
\label{app:tasks}
\begin{table}[!htbp]
  \centering
  \caption{\textbf{Tasks.}}
  \label{tab:tasks}
  \footnotesize
  \begin{tabular}{@{}l >{\raggedright\arraybackslash}p{0.78\linewidth}@{}}
    \toprule
    RoboDojo (25) & \task{align\_blocks}, \task{build\_tower}, \task{deposit\_coin}, \task{fill\_egg\_holder}, \task{fill\_pen\_holder}, \task{hang\_mugs\_random}, \task{insert\_key}, \task{insert\_tubes}, \task{make\_kong}, \task{make\_toast\_random}, \task{match\_and\_pick\_from\_conveyor}, \task{organize\_table}, \task{pack\_objects\_into\_box}, \task{pick\_from\_conveyor\_by\_image}, \task{play\_Xylophone}, \task{play\_stacking\_toy}, \task{plug\_in\_charger}, \task{pour\_by\_language}, \task{pour\_liquid\_into\_cup}, \task{press\_by\_number}, \task{stack\_bowls\_random}, \task{store\_laptop\_and\_headphones}, \task{store\_tools\_in\_toolbox}, \task{swap\_blocks}, \task{sweep\_blocks\_random} \\
    LIBERO-PRO (6) & \texttt{libero\_10\_swap} tasks 5, 6, 9; \texttt{libero\_goal\_swap} tasks 1, 5, 9 \\
    RoboCasa365 (10) & \texttt{OpenCabinet}, \texttt{PickPlaceDrawerToCounter}, \texttt{TurnOffStove}, \texttt{HeatKebabSandwich}, \texttt{LoadDishwasher}, \texttt{MakeIceLemonade}, \texttt{SeparateFreezerRack}, \texttt{SteamInMicrowave}, \texttt{StirVegetables}, \texttt{WaffleReheat} \\
    RoboCasa365 sources (10) & \texttt{CloseBlenderLid}, \texttt{CloseToasterOvenDoor}, \texttt{CoffeeSetupMug}, \texttt{OpenStandMixerHead}, \texttt{TurnOnSinkFaucet}, \texttt{ArrangeBreadBasket}, \texttt{GatherTableware}, \texttt{PackIdenticalLunches}, \texttt{SearingMeat}, \texttt{StoreLeftoversInBowl} \\
    \bottomrule
  \end{tabular}
\end{table}

\FloatBarrier
\section{Comparing Criteria with Instructions in RoboDojo}
\label{app:audits}
\label{app:audit}
\paragraph{Criteria mechanism.} RoboDojo criteria are sequences of stages: \texttt{check} stages in order,
\texttt{final\_check} at termination, and \texttt{trigger\_check}, evaluated only at the step its condition becomes true;
\texttt{query} requires exact counts and fails the episode for good once a count is exceeded; conditions nest with
alternating AND and OR. Of the \V{AUDIT_TASKS} tasks, \V{AUDIT_HOME_TASKS} require both arms to return within 15\,cm and
20\textdegree{} of home, while only \V{AUDIT_HOME_INSTR} instructions ask for a reset.

\paragraph{Method.} For each task we compare \texttt{gen\_instruction}, \texttt{run\_reward}, \texttt{get\_score} and the
configuration, grade the severity as high, medium or low, and confirm by hand; the review is mainly a static reading of the
code.

\paragraph{Types.} The type names are ours.
\begin{itemize}
  \item \emph{Hidden rule}: the criteria verify conditions the instruction does not state (order, counts, relative
    position), e.g.\ the button protocol of \task{press\_by\_number} or the tile to draw in \task{make\_kong}.
  \item \emph{Hard threshold}: the tolerance is stricter than clearly satisfying the instruction requires, e.g.\ 97\% of
    the liquid in \task{pour\_liquid\_into\_cup}.
  \item \emph{Timing, event triggers and irreversibility}: verification at one step only, or permanent failure once an event
    happens, e.g.\ pouring judged only when the bottle returns upright; a third press in \task{press\_by\_number}.
  \item \emph{Criteria defect}: the criteria code or an asset annotation is wrong, e.g.\ mug 14 in
    \task{hang\_mugs\_random}, the bug in \task{solve\_equation}.
  \item \emph{Extra end state}: an end state the instruction does not mention, e.g.\ in \task{make\_toast} the two
    remaining slices must stay on the bread shelf.
  \item \emph{Loose criteria}: trajectories that skip the action the instruction asks for are accepted, e.g.\
    \task{sweep\_blocks} accepts blocks picked up without the broom.
  \item \emph{Contradiction}: the criteria require the opposite of the instruction, e.g.\ in
    \task{classify\_objects\_by\_language} the categories are named after asset folders, so a chocolate bunny counts
    as a wooden toy.
\end{itemize}

\paragraph{Counts.} 23 tasks are consistent and 19 inconsistent (hidden rule 9, hidden rule and hard threshold 5, hard
threshold 3, contradiction 1, defect 1); with \task{solve\_equation}, whose criteria have a bug, \V{AUDIT_CL_TASKS} tasks
and \V{AUDIT_CL_VARIANTS} variants are criteria-limited. By type, at medium or high severity: hidden rule
\V{AUDIT_HIDDEN}, hard threshold \V{AUDIT_THRESH}, timing and events \V{AUDIT_EVENT}, defect \V{AUDIT_DEFECT}, extra end
state \V{AUDIT_ENDSTATE}, loose criteria \V{AUDIT_LOOSE}, contradiction
\V{AUDIT_CONTRA}. Other typical cases are \task{make\_kong} (a tile must be drawn) and \task{build\_tower} (only one
configuration is accepted).

\paragraph{Published rules.} The task descriptions and scoring tables on the RoboDojo website publish part of these
rules, for example the button order of \task{press\_by\_number}; the recipes of EmbodiedEvo and HexaShell come from
there~\citep{li2026embodiedevo,gao2026hexashell}. \Cref{tab:audit-matrix} gives the tasks by type. A task is
criteria-limited only if the manual review confirms that its criteria and its instruction disagree; the table also
lists the findings for tasks that the review kept as faithful.

\begin{table}[!htbp]
  \centering
  \caption{\textbf{Tasks by type.} \ding{108}: high severity; \ding{109}: medium; blank: low or none. The last
    column marks the 33 tasks whose criteria require both arms to return home (not counted as a type). Total: tasks at
    medium or high severity (high in parentheses). $^{r}$: reviewed together with its \task{\_random} variant.}
  \label{tab:audit-matrix}
  \footnotesize
  \setlength{\tabcolsep}{4pt}
  \renewcommand{\arraystretch}{0.92}
  \begin{tabular}{@{}l *{7}{c} c@{}}
    \toprule
    Task & \makecell{Hidden\\rule} & \makecell{Hard\\threshold} & \makecell{Timing/\\irrev.} & \makecell{Criteria\\defect} & \makecell{Extra\\end state} & \makecell{Loose\\criteria} & \makecell{Contra-\\diction} & \makecell{Arms\\home} \\
    \midrule
    \task{align\_blocks} &  & \ding{109} & \ding{109} &  &  &  &  & \checkmark \\
    \task{arrange\_largest\_number}$^{r}$ &  &  &  &  &  &  &  & \checkmark \\
    \task{build\_tower} & \ding{108} & \ding{109} &  & \ding{109} &  &  &  & \checkmark \\
    \task{classify\_objects} &  &  &  &  &  &  &  & \checkmark \\
    \task{classify\_objects\_by\_language} &  &  &  &  &  &  & \ding{109} & \checkmark \\
    \task{cover\_blocks} &  &  & \ding{109} &  & \ding{109} &  &  & \checkmark \\
    \task{deposit\_coin} &  & \ding{109} &  &  &  &  &  & \checkmark \\
    \task{fasten\_screws} &  & \ding{109} &  &  &  & \ding{109} & \ding{109} & \checkmark \\
    \task{fill\_egg\_holder} &  &  &  &  &  &  &  & \checkmark \\
    \task{fill\_pen\_holder} & \ding{109} &  &  &  &  &  &  & \checkmark \\
    \task{fold\_clothes}$^{r}$ & \ding{109} &  & \ding{109} &  &  &  &  & \checkmark \\
    \task{general\_pickup} &  &  &  &  &  &  &  &  \\
    \task{hang\_mugs}$^{r}$ &  &  &  & \ding{108} &  &  &  & \checkmark \\
    \task{imitate\_sorting\_sequence} &  &  & \ding{109} &  &  &  &  & \checkmark \\
    \task{insert\_key} & \ding{109} &  &  &  &  &  &  &  \\
    \task{insert\_tubes} &  &  &  &  &  &  &  & \checkmark \\
    \task{make\_kong} & \ding{108} &  & \ding{109} &  & \ding{109} &  &  &  \\
    \task{make\_toast}$^{r}$ &  &  &  & \ding{109} & \ding{109} &  &  & \checkmark \\
    \task{match\_and\_pick\_from\_conveyor} &  &  &  &  &  & \ding{109} &  &  \\
    \task{organize\_table} & \ding{109} & \ding{109} &  &  &  &  &  & \checkmark \\
    \task{pack\_objects\_into\_box}$^{r}$ & \ding{109} & \ding{109} &  &  &  &  &  & \checkmark \\
    \task{pick\_from\_conveyor\_by\_image} &  &  &  &  &  &  &  &  \\
    \task{play\_stacking\_toy} & \ding{109} & \ding{109} &  &  &  &  &  & \checkmark \\
    \task{play\_tic\_tac\_toe} & \ding{109} &  & \ding{109} & \ding{109} &  &  &  &  \\
    \task{play\_Xylophone} &  &  &  &  &  & \ding{109} &  &  \\
    \task{plug\_in\_charger} &  &  &  &  &  &  &  & \checkmark \\
    \task{pour\_balls\_into\_vase} &  &  &  & \ding{109} & \ding{109} &  &  & \checkmark \\
    \task{pour\_by\_language} &  & \ding{109} & \ding{109} & \ding{109} &  &  &  &  \\
    \task{pour\_liquid\_into\_cup}$^{r}$ & \ding{109} & \ding{109} & \ding{109} &  &  &  &  &  \\
    \task{press\_by\_number} & \ding{109} &  & \ding{108} & \ding{108} &  &  & \ding{108} & \checkmark \\
    \task{push\_T}$^{r}$ &  & \ding{109} &  &  &  &  &  & \checkmark \\
    \task{put\_bottles\_into\_dustbin} &  &  &  &  &  &  &  & \checkmark \\
    \task{solve\_equation} &  &  &  & \ding{109} &  &  &  & \checkmark \\
    \task{sort\_nesting\_dolls\_by\_size}$^{r}$ &  & \ding{109} &  &  &  &  &  & \checkmark \\
    \task{stack\_blocks}$^{r}$ &  &  &  &  &  &  &  & \checkmark \\
    \task{stack\_blocks\_by\_language} &  &  &  &  &  &  &  & \checkmark \\
    \task{stack\_bowls}$^{r}$ &  &  &  &  &  &  &  & \checkmark \\
    \task{store\_laptop\_and\_headphones}$^{r}$ & \ding{109} &  &  &  &  &  &  & \checkmark \\
    \task{store\_tools\_in\_toolbox} &  & \ding{109} &  &  &  &  &  & \checkmark \\
    \task{swap\_blocks} &  &  & \ding{109} & \ding{108} &  &  &  & \checkmark \\
    \task{swap\_T} & \ding{109} & \ding{108} &  &  &  &  &  & \checkmark \\
    \task{sweep\_blocks}$^{r}$ &  &  &  &  & \ding{109} & \ding{109} &  & \checkmark \\
    \midrule
    Total & 13 (2) & 13 (1) & 10 (1) & 9 (3) & 5 (0) & 4 (0) & 3 (1) & 33 \\
    \bottomrule
  \end{tabular}
\end{table}

\FloatBarrier
\section{Instruction-faithful Criteria}
\label{app:criteria}
\paragraph{Rules.} (1)~Change only \texttt{run\_reward} and \texttt{get\_score}, never the instruction. (2)~Keep the
return of both arms. (3)~Remove or relax hidden rules. (4)~Fix defects. (5)~Tighten loose criteria where the simulator
state detects the difference reliably. (6)~Relax hard thresholds to what still clearly satisfies the instruction, listing
old and new values. (7)~Correct asset annotations in memory only. (8)~Leave the original repository untouched. New
criteria primitives check distance and orientation on the same feature, containment in an elliptic cylinder, and
connected towers.

\paragraph{Validation.} Offline unit tests; shadow judging and reruns on seed~0; on seed~2 both criteria judge the same
episodes independently (an episode ends only when both have a verdict), and every flipped episode was checked by hand.

\paragraph{Decisions of our own.} ``Tighten'' in \task{fasten\_screws} means a nut sinking at least 35\,mm. We also decided what counts as a tower in
\task{build\_tower} and which edges of the laptop may face down in \task{store\_laptop\_and\_headphones}.
\Cref{tab:criteria-changes} lists every change.

\begingroup
\scriptsize
\renewcommand{\arraystretch}{1.1}
\setlength{\LTcapwidth}{\linewidth}
\begin{xltabular}{\linewidth}{@{}>{\raggedright\arraybackslash}p{2.5cm} c >{\raggedright\arraybackslash}X >{\raggedright\arraybackslash}X@{}}
  \caption{\textbf{Changes from $\Corig$ to $\Cfaith$.} Type: D, defect fixed; H, hidden rule removed; R, hard
    threshold relaxed; T, loose criterion tightened. A \task{\_random} variant receives the same change unless marked.
    Conditions not listed, including the return of both arms, are unchanged. Asset annotations are corrected in memory
    only. Liquid thresholds are shares of the poured liquid in the target and left in the bottle.}
  \label{tab:criteria-changes}\\
    \toprule
    Task & Type & $\Corig$ & $\Cfaith$ \\
    \midrule
  \endfirsthead
    \multicolumn{4}{@{}l}{\textit{Table~\ref{tab:criteria-changes}, continued}} \\
    \toprule
    Task & Type & $\Corig$ & $\Cfaith$ \\
    \midrule
  \endhead
    \bottomrule
  \endlastfoot
    \task{build\_tower} & H & one accepted 6-level configuration: given boards on given pillar pairs, pillars grouped
      $\{1,5\}/\{2,6\}$, all 8 parts tilted $<5^\circ$, top two levels centred within 2.3\,cm, $20^\circ$ between boards &
      one connected stack of all 8 parts, each on the table or on another part, longest support chain $\ge 4$ levels, all
      parts at rest ($<2$\,mm per step) \\
    & D & ``A on B'' compares heights only & bottom within 1.5\,cm of the top below and overlapping in $xy$ \\
    & -- & score tiers of the fixed configuration & $\ge 3$ parts in 2 levels: 10; $\ge 5$ in 3 levels: 30; full: 100 \\
    \midrule
    \task{fasten\_screws} & T & nut sunk $\ge 9$\,mm on its bolt (``tighten'') & $\ge 35$\,mm (seated $\approx 38$\,mm) \\
    & R & nut coaxial with bolt within 1\,mm & within 4\,mm \\
    \midrule
    \task{fill\_egg\_holder} & D & ``in holder'' by the oriented box, which reaches 5.5\,cm behind the body & also within
      5.5\,cm of the holder centre in its own frame \\
    \midrule
    \task{fill\_pen\_holder} & H & every pen tip down (4 checks); $xy$ of the pen's check end within 3.5\,cm of the axis &
      no tip direction; pen centre within 3.5\,cm of the axis (depth 3.5\,cm kept) \\
    & T & holder never verified (``hold \ldots\ then put it back down'') & holder lifted $>2$\,cm at some step and back within
      1.5\,cm of its start height at the end \\
    \midrule
    \task{hang\_mugs} & D & orientation compared with hook~0 only, distance with any hook (hooks 2 and 3 never pass) &
      distance $<4.5$\,cm, $z$ axis $<60^\circ$ and $x$ axis $\pm 60^\circ$ on the same hook (values kept) \\
    & D & mug/00014 handle frame~0 annotated as identity rotation & orientation of the other mugs' handles \\
    \midrule
    \task{insert\_key} & H & key turned clockwise by $30$--$90^\circ$ & turned $>30^\circ$ in either direction \\
    & T & handover not verified & both hands have held the key $>2$\,cm off the table (gripper reading 0.01--0.6) before
      insertion; new score tier \\
    \midrule
    \task{make\_kong} & H & draw the replacement tile, stand it at $(0.319, -0.15)$ within 1.5\,cm and $7^\circ$, grippers
      open; tile untouched before the kong & removed \\
    & R & other hand tiles standing within $7^\circ$ & within $20^\circ$ \\
    & T & standing pair \texttt{mahjong4\_0/4\_1} not verified & must stay standing (upstream fix 21b8d23) \\
    \midrule
    \task{make\_toast} & D & each slot accepts any slice (one slice can fill both) & two different slices, one per slot \\
    & D & \emph{random:} slot corners of toaster/00004 mixed up, one corner of toaster/00002 slot~2 wrong & corrected from the
      USD geometry (3\,mm margin) \\
    & H & exactly 2 slices left on the rack; all 4 slices upright within $30^\circ$ & removed \\
    & T & order of bread and lever not verified & stage 1: both slices in, lever $<85\%$; stage 2: lever $>85\%$, arms back \\
    \midrule
    \task{organize\_table} & R & figurine on stand: stand centre within $\{4.5, 2.5, 4.5, 2.0, 3.2\}$\,cm of the base
      (models 5, 6, 7, 8, 11) & within the base's contact radius, $\{7.15, 4.27, 6.78, 2.41, 4.83\}$\,cm \\
    & H & mouse local $y$ within $45^\circ$ of world $+x$ (signed) & mouse upright within $15^\circ$, heading free \\
    & H & keyboard local $+y$ within $45^\circ$ of world $+y$ & $+y$ or $-y$ \\
    & T & alarm clock: height only & also over the drawer footprint \\
    & D & at-rest checks in \texttt{get\_score} void (state shared with \texttt{run\_reward}) & dropped from
      \texttt{get\_score} \\
    \midrule
    \task{pack\_objects\_into\_box} & H & object long axis within $45^\circ$ of the box's long side; box heading within
      $40^\circ$ & removed (front within $60^\circ$ of $-x$, box upright $15^\circ$ kept) \\
    \midrule
    \task{play\_tic\_tac\_toe} & D & piece within 1.5\,cm of the cell centre, while the opponent's moves count at 1.8\,cm &
      1.8\,cm (\texttt{run\_reward} and \texttt{get\_score}) \\
    \midrule
    \task{pour\_balls\_into\_vase} & D & ``in vase'': $xy$ hull of the oriented box and above the bottom (a tilted
      27$\times$24$\times$28\,cm box for vase/00002) & inside the elliptic cylinder inscribed in the body, between bottom and
      rim \\
    & H & cup upright within $15^\circ$ after pouring & removed \\
    \midrule
    \task{pour\_by\_language} & D & order verified only when either arm reaches its origin & verified at every step \\
    & H & each bowl judged only at the step its bottle is back within $30^\circ$ of upright & judged at every step, bottle
      may stay tilted \\
    & R & bowl $\ge 95\%$, bottle $\le 10\%$ & bowl $\ge 90\%$, bottle $\le 15\%$ \\
    & T & reset of the arms (``Then reset the robot arm'') not verified & all three bowls poured with both arms at the origin \\
    \midrule
    \task{pour\_liquid\_into\_cup} & H & judged only at the step the bottle is back within $30^\circ$ of upright & judged at
      every step \\
    & R & cup $\ge 97\%$ (bottle $\le 15\%$) & cup $\ge 90\%$ (bottle $\le 15\%$) \\
    \midrule
    \task{press\_by\_number} & D & press counted when the cap goes from $>0.95$ to $<0.5$ of its travel, released at $>0.9$
      (a released cap rests near 0.75) & from $>0.65$ to $<0.5$, released at $>0.65$ \\
    & H & left red $\times n_0$, blue, middle red $\times n_1$, blue; blue exactly twice (contradicts ``then press the blue
      button to confirm'') & each red button exactly its card's number, in any order, then blue once; any extra press or
      an early blue fails \\
    \midrule
    \task{solve\_equation} & D & position, $x$ and $y$ orientation each satisfied by any candidate piece & all three by the
      same piece (as upstream 29910f3; 1.8\,cm and $45^\circ$ kept) \\
    \midrule
    \task{store\_laptop\_and\_headphones} & H & headset: whole orientation within $30^\circ$ of the annotation &
      headset $z$ axis up within $30^\circ$, heading free (headband within 5\,cm kept) \\
    & H & laptop vertical: hinge up within $10^\circ$ & any edge down ($\pm x$ or $\pm y$ up within $10^\circ$) \\
    & H & in the stand: the front-edge midpoint of the base & any of the four edge midpoints \\
    \midrule
    \task{store\_tools\_in\_toolbox} & R & tool orientation within $10^\circ$ of its slot & within $15^\circ$; also in
      \texttt{get\_score} \\
    \midrule
    \task{swap\_blocks} & D & button counting as in \task{press\_by\_number} ($0.95/0.9$) & $0.65$ \\
    & H & block lifted 3\,cm before each move (3 stages) & removed \\
    & T & blocks may leave the swapped positions after the last press & still swapped at the end \\
    \midrule
    \task{sweep\_blocks} & T & broom never verified (``Pick up the broom'') & lifted 5\,cm above its start at some step \\
    & R & dustpan left of the broom by $>0.1$\,m & by $>0$ \\
    & R & dustpan tilted $<5^\circ$ & $<20^\circ$ \\
    & H & dustpan not lifted $>1$\,cm & removed \\
    & T & blocks over the pan in $xy$ (a block under the pan passes) & each block inside the pan in the dustpan's frame \\
\end{xltabular}
\endgroup

\section{Online Dual-judged Results}
\label{app:dual}
\Cref{tab:dual16} gives the per-task counts of \cref{sec:exp-rejudge}.
\begin{table}[!htbp]
  \centering
  \caption{\textbf{The same episodes judged by both criteria, per task} (seed~0, 20 layouts per task and library).
    $n_{11}$: success under $\Corig$ and $\Cfaith$; $n_{10}$: only under $\Corig$; $n_{01}$: only under $\Cfaith$;
    $n_{00}$: under neither. Libraries of \cref{sec:exp-orig-main}.}
  \label{tab:dual16}
  \footnotesize
  \setlength{\tabcolsep}{4pt}
  \begin{tabular}{@{}l *{4}{c} @{\hspace{12pt}} *{4}{c}@{}}
    \toprule
    & \multicolumn{4}{c}{\wCE} & \multicolumn{4}{c}{\woCE} \\
    \cmidrule(lr){2-5} \cmidrule(l){6-9}
    Task & $n_{11}$ & $n_{10}$ & $n_{01}$ & $n_{00}$ & $n_{11}$ & $n_{10}$ & $n_{01}$ & $n_{00}$ \\
    \midrule
    \task{build\_tower} & 20 & 0 & 0 & 0 & 20 & 0 & 0 & 0 \\
    \task{fill\_egg\_holder} & 4 & 0 & 0 & 16 & 5 & 0 & 0 & 15 \\
    \task{fill\_pen\_holder} & 11 & 0 & 0 & 9 & 2 & 1 & 1 & 16 \\
    \task{hang\_mugs\_random} & 7 & 0 & 1 & 12 & 1 & 0 & 2 & 17 \\
    \task{insert\_key} & 0 & 0 & 0 & 20 & 0 & 0 & 0 & 20 \\
    \task{make\_kong} & 20 & 0 & 0 & 0 & 0 & 0 & 20 & 0 \\
    \task{make\_toast\_random} & 7 & 2 & 0 & 11 & 1 & 1 & 0 & 18 \\
    \task{organize\_table} & 8 & 0 & 2 & 10 & 12 & 0 & 2 & 6 \\
    \task{pack\_objects\_into\_box} & 15 & 0 & 0 & 5 & 10 & 0 & 2 & 8 \\
    \task{pour\_by\_language} & 19 & 0 & 0 & 1 & 19 & 0 & 0 & 1 \\
    \task{pour\_liquid\_into\_cup} & 15 & 0 & 3 & 2 & 10 & 0 & 2 & 8 \\
    \task{press\_by\_number} & 0 & 20 & 0 & 0 & 0 & 0 & 19 & 1 \\
    \task{store\_laptop\_and\_headphones} & 0 & 0 & 0 & 20 & 0 & 0 & 2 & 18 \\
    \task{store\_tools\_in\_toolbox} & 0 & 0 & 0 & 20 & 0 & 0 & 0 & 20 \\
    \task{swap\_blocks} & 0 & 0 & 0 & 20 & 0 & 0 & 14 & 6 \\
    \task{sweep\_blocks\_random} & 12 & 5 & 0 & 3 & 7 & 0 & 0 & 13 \\
    \midrule
    Total (320 each) & 138 & 27 & 6 & 149 & 87 & 2 & 64 & 167 \\
    Successes, $\Corig$ / $\Cfaith$ & \multicolumn{4}{c}{165 / 144} & \multicolumn{4}{c}{89 / 151} \\
    \bottomrule
  \end{tabular}
\end{table}

\section{Reading the Instruction-faithful Criteria}
\label{app:readfaith}
\paragraph{Configuration.} Ten of the 16 rewritten criteria-limited tasks, two per RoboDojo dimension, learned with
Criteria Exposure to $\Cfaith$ for five rounds: round~0 runs the empty library, and rounds~1--5 make one revision
each. Each round runs the five development layouts 0--4 of seed~1, 300 episodes in all; a
revision with at least two fewer successes than the best earlier round is rolled back, and no test runs during learning.
Each episode is judged by $\Cfaith$, which drives the loop, and by $\Corig$ alongside, which is only logged. The actor
is GPT-6 Astra with high reasoning effort; the learner is Claude Code with Claude Opus~5.5 and high reasoning
effort, with 120 minutes and at most eight replays per session. The learner could read the development episodes of the parent revision (code cells, model
turns, primitive calls, frames, prompt, verdicts and scores), the findings of earlier rounds, the PyRUA-Lean API sources,
and the $\Cfaith$ criteria code of its own task. It was not given the $\Corig$ criteria code, other tasks' criteria or
runs, the assets, or the sources and notes of the rewrite. The seed-0 test
takes the library active after round~5 (for \task{fill\_pen\_holder} that of round~1, since rounds~2--5 were rolled
back) and runs layouts 0--19 of seed~0 with the same actor, each episode judged by $\Cfaith$ and by $\Corig$.

\begin{table}[!htbp]
  \centering
  \caption{\textbf{The library learned by reading $\Cfaith$, judged by both criteria.} Seed-0 test of the final
    library on 20 layouts per task, the same episodes judged by $\Cfaith$, which the learner read, and by $\Corig$.
    Dim.: RoboDojo dimension
    (Generalization, Precision, Long-horizon, Memory, Open).}
  \label{tab:readfaith}
  \footnotesize
  \begin{tabular}{@{}l l c c@{}}
    \toprule
    Task & Dim. & $\Cfaith$ & $\Corig$ \\
    \midrule
    \task{sweep\_blocks\_random} & Gen. & 20/20 & 20/20 \\
    \task{hang\_mugs\_random} & Gen. & 4/20 & 2/20 \\
    \task{insert\_key} & Prec. & 0/20 & 0/20 \\
    \task{build\_tower} & Prec. & 20/20 & 0/20 \\
    \task{make\_kong} & Long & 20/20 & 0/20 \\
    \task{fill\_pen\_holder} & Long & 7/20 & 5/20 \\
    \task{swap\_blocks} & Mem. & 20/20 & 0/20 \\
    \task{press\_by\_number} & Mem. & 20/20 & 0/20 \\
    \task{store\_tools\_in\_toolbox} & Open & 0/20 & 0/20 \\
    \task{pour\_by\_language} & Open & 19/20 & 0/20 \\
    \midrule
    Total & & 130/200 & 27/200 \\
    \bottomrule
  \end{tabular}
\end{table}

\FloatBarrier
\section{What Exposure Writes into the Library}
\label{app:writes}

\paragraph{Setup.} We classify the memory entries of the ten source libraries of the RoboDojo transfer
(\cref{app:transfer}), for both settings, by whether they cite the criteria code, and compare entries that the two settings wrote from the same source failures.

\paragraph{Results.} Two patterns emerge. First, an exposed library writes its experience in the language of the
criteria: \V{CARDS_ON_CRIT} of the \V{CARDS_ON_N} exposed entries cite the criteria code (\V{CARDS_ON_CHARS}\% of the
characters), and none of the \V{CARDS_OFF_N} success-only entries does. \Cref{fig:writes-quotes} shows three exposed entries: the
grip-and-pull of the button cap in \task{swap\_blocks} and the ban on sweeping in \task{sweep\_blocks\_random}, which
cite the criteria code, and the hidden protocol of \task{press\_by\_number}, which restates the order of the checks in
words.

\begin{figure}[!htbp]
  \centering
  \footnotesize
  \begin{minipage}{\linewidth}
    \textbf{\task{press\_by\_number}, \texttt{press\_by\_number-manipulate-001}} (learner \wCE)
\begin{lstlisting}[style=card]
do: Press in exactly this order: LEFT red xN_left (left arm) -> BLUE once (right arm) ->
MIDDLE red xN_middle (right arm) -> BLUE once more -> ... Blue is pressed exactly TWICE, never more. ...
evidence: task checks in order: repeat(button0 below 0.5 / above 0.9) x num0; button2 below 0.5; ...
\end{lstlisting}
    \vspace{4pt}
    \textbf{\task{swap\_blocks}, \texttt{swap\_blocks-verify-001}} (learner \wCE)
\begin{lstlisting}[style=card]
do: the benchmark judges after every action, in order: lift block 1 -> block 1 on the empty mat ->
cap pressed below 50% of its travel -> cap seen back ABOVE 90% -> ... a push alone ... does not
bring the cap back above 90%: the fingers must grip the cap and pull it up.
\end{lstlisting}
    \vspace{4pt}
    \textbf{\task{sweep\_blocks\_random}, \texttt{sweep\_blocks\_random-manipulate-001}} (learner \wCE)
\begin{lstlisting}[style=card]
do: do NOT pick up, hand over or sweep with the broom. The benchmark only checks, in one action:
every object's centre inside the dustpan's bin, dustpan not lifted and upright, broom's x >
dustpan's x + 0.10 m, both arms home. Carry the objects in by hand: ...
\end{lstlisting}
  \end{minipage}
  \caption{\textbf{Exposed entries are written in the language of the criteria.} Excerpts of three memory entries
    of the exposed source libraries, verbatim (``\dots'' marks elisions). The first fixes an order of presses that the
    instruction does not state, with the verification order of the criteria as evidence; the second turns a verified condition
    (``above 90\%'') into a grip-and-pull of the button cap; the third tells the actor not to use the broom the
    instruction names (``pick up the broom, hand it over to the right hand''), because the verifier never looks at it. All three
    are marked \texttt{dev\_verified}; the on-10 distillation dropped all three.}
  \label{fig:writes-quotes}
\end{figure}
 Such entries fail under $\Cfaith$ and have to be
removed for transfer: distilling on-10 dropped \V{CARDS_DROPPED} entries on ``what the benchmark checks''. Second, a
learner that cannot see the criteria has to find a physical cause. From the same source failures, in which the gripper
closes on nothing, the success-only learner traced the cause to a half-open gripper whose fingers \texttt{KEEP} moves
close again, and wrote ``pass \texttt{OPEN} on every move of an empty gripper''; the exposed learner recorded the
symptom and kept a formula that in effect half-opens the gripper (\cref{fig:case}). Not all exposed entries concern the
verifier, though: some carry information the instruction should state, such as the pen orientation in
\task{fill\_pen\_holder}.

\FloatBarrier
\section{Library Transfer}
\label{app:transfer}
To reuse libraries across tasks, we distill the libraries of ten source tasks into one shared library and apply it to
disjoint target tasks; entries of exposed libraries that are tied to the criteria are removed during distillation.
Across benchmarks, the libraries are applied on LIBERO-PRO and RoboCasa365 with primitives aligned to those of RoboDojo
(Appendix~\ref{app:align}); within RoboCasa365 we also run both settings of the loop and distill one library from each,
as a reference for in-benchmark experience.
In RoboDojo, the success-only and exposed libraries off-10 and on-10 take ten learned tasks as sources and are
applied to ten target tasks that no source covers (\cref{tab:transfer}); off-rc10 and on-rc10 are distilled within
RoboCasa365 from both settings of the loop. The case study is in Appendix~\ref{app:case}.

\paragraph{Brief.} The distiller receives, for each source task, the library as the actor read it in the last
evaluated revision (retired entries removed), the learner's last diagnosis and the development successes per round,
and the robot API source; it does not see the target tasks and is told not to look for them. It writes memory
entries of at most 12{,}000 characters in the source format (\texttt{stage}, \texttt{symptom}, \texttt{applies},
\texttt{do}, \texttt{evidence}) and at most two skill modules of 15{,}000 characters. Every entry must state a general
rule, name no task, object, dimension or goal configuration in \texttt{do}, and cite evidence from at least two source
tasks, or from one task where it decided between failing and succeeding rounds, flagged as a single-task rule; a
separate notes file, not given to the actor, records the source of every entry and what was left out. The two briefs
differ in one passage on the learners: ``they never saw the benchmark's judging code'' (\woCE) against ``A rule that
only describes how one task is judged (its thresholds, its check order) does not transfer; keep what is about
perception and manipulation'' (\wCE).

\paragraph{Process.} Each library is distilled in one session by Claude Code with Claude Opus 5.5 (effort high).
Before
evaluation, every library is checked for the character limits, the entry fields, source task names, and that each
library function it names exists and parses; skills are exercised against a stub robot with the real API signatures,
not in simulation.

\paragraph{Results.} \Cref{tab:case-pertask} gives the ten target tasks. Off-10 leads on-10 by 9 episodes, and
most of the difference sits in \task{insert\_tubes} and \task{solve\_equation} (Appendix~\ref{app:case}).

\FloatBarrier
\section{Case Study: What Carries Over within RoboDojo}
\label{app:case}
\label{sec:exp-why}

\paragraph{Setup.} We follow the ten-source distillation of \cref{sec:exp-tasks}, whose success-only and exposed
libraries we call off-10 and on-10, on the ten target tasks. For each episode we count the gripper arguments passed to
primitives, the moves made to re-measure with a wrist camera, and the use of the action budget.

\begin{figure}[!htbp]
  \centering
  \begin{minipage}[t]{0.49\linewidth}
    \footnotesize\textbf{Source, \woCE: \task{make\_toast\_random}, diagnosis}
\begin{lstlisting}[style=card]
**Right fingers arrive shut at the hand-off.**
The host's KEEP sends "close" whenever the
measured opening reads below the last command
... 18 right closes began at opening 0.0, and
**none held**.
\end{lstlisting}
    \vspace{3pt}
    \footnotesize\textbf{Distilled, off-10: \texttt{general-grasp-004}}
\begin{lstlisting}[style=card]
do: the fingers must really be open: pass
gripper=robo.OPEN on every move of an empty
gripper (KEEP re-closes fingers that read below
the last command) and check
state().arm(arm).gripper >= 0.9 before
descending. ...
\end{lstlisting}
  \end{minipage}\hfill
  \begin{minipage}[t]{0.49\linewidth}
    \footnotesize\textbf{Source, \wCE: \task{make\_toast\_random}, \texttt{grasp-002}}
\begin{lstlisting}[style=card]
symptom: grasps from straight above missed or
tipped neighbours; closes 3-6 cm short held
nothing and left rack slices leaning. ...
\end{lstlisting}
    \vspace{3pt}
    \footnotesize\textbf{Distilled, on-10: \texttt{grasp-004}}
\begin{lstlisting}[style=card]
do: Open fully unless you must fit between
close neighbours (set_gripper(arm, robo.OPEN,
opening=w): 0.002+0.109*w m, a bit over the
object; ...). Descent stopped > ~1.2 cm above
target or blocked: do NOT close; rise,
re-plan ...
\end{lstlisting}
  \end{minipage}

  \vspace{6pt}
  \includegraphics[width=\linewidth]{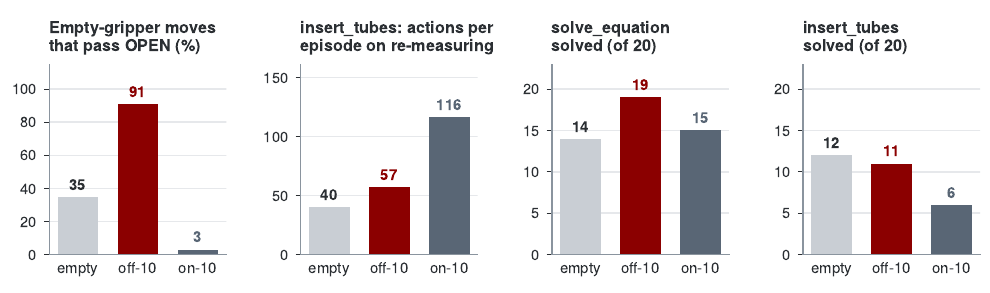}
  \caption{\textbf{Two library entries with the same number, from the same source failures.} Top: the source entry
    each learner wrote on \task{make\_toast\_random}, where closes held nothing, and the grasp entry distilled from it.
    The success-only learner traces the empty closes to \texttt{KEEP} moves and off-10 generalises the cause; the
    exposed learner records the symptom and on-10 keeps a formula that opens the gripper only part way. Verbatim;
    ``\dots'' marks elisions. Bottom: what the actor does with them on the ten target tasks, for the empty library, off-10 and on-10. Closes after a
    half-open gripper and \texttt{KEEP} moves grasp nothing in \V{CS_EMPTY_HALF}\% of cases, closes after \texttt{OPEN}
    moves in \V{CS_EMPTY_OPEN}\%.}
  \label{fig:case}
\end{figure}

\paragraph{Results.} \textbf{General rules carry over and single-task precision does not.} Both libraries hold a grasp
entry with the same number, drawn from the same source failures (\cref{fig:case}). The success-only learner traced the
failures to their cause, fingers that \texttt{KEEP} moves close again, and wrote a general rule; the exposed learner kept
a formula that half-opens the gripper. The actor follows each grasp entry, passing \texttt{OPEN} on \V{CS_OPEN_V0}\%,
\V{CS_OPEN_OFF}\% and \V{CS_OPEN_ON}\% of empty-gripper moves with the empty library, off-10 and on-10, and on
\task{solve\_equation}, which needs a thin disc picked up, off-10 solves \V{CS_SOLVE_OFF}/20 layouts against
\V{CS_SOLVE_V0} and \V{CS_SOLVE_ON}. On-10 instead asks to re-measure to 1\,mm before most actions; on
\task{insert\_tubes} it spends \V{CS_HOVER_ON} of 500 actions per episode on re-measuring (\V{CS_HOVER_V0} empty,
\V{CS_HOVER_OFF} off-10) and solves \V{CS_TUBES_ON}/20 against \V{CS_TUBES_V0} and \V{CS_TUBES_OFF}.

\begin{table}[!htbp]
  \centering
  \caption{\textbf{The ten target tasks with the empty library, off-10 and on-10.} Successes under $\Corig$ on
    layouts 0--19 of seed 0. OPEN: share of
    \texttt{move\_pose} calls of an empty gripper that pass \texttt{gripper=OPEN} among those passing \texttt{OPEN}
    or \texttt{KEEP}.
    Re-measure: actions per episode spent on moves that are directly followed by a wrist-camera perception call.
    OPEN and re-measure use only the first run of each layout; episodes run again are excluded, so off-10 and on-10
    have the number in the last column and the empty library 20 per task. Values marked $^\dagger$ rest on fewer than ten episodes.}
  \label{tab:case-pertask}
  \footnotesize
  \setlength{\tabcolsep}{3.5pt}
  \begin{tabular}{@{}l r ccc ccc ccc c@{}}
    \toprule
    & & \multicolumn{3}{c}{Successes} & \multicolumn{3}{c}{OPEN (\%)} & \multicolumn{3}{c}{Re-measure actions} & Episodes \\
    \cmidrule(lr){3-5}\cmidrule(lr){6-8}\cmidrule(lr){9-11}
    Task & Budget & $\Lempty$ & off-10 & on-10 & $\Lempty$ & off-10 & on-10 & $\Lempty$ & off-10 & on-10 & off/on \\
    \midrule
    \task{fasten\_screws}               & 1900 & 11 & 12 & 12    & 38 & 89$^\dagger$  & 0$^\dagger$  & 173 & 230$^\dagger$ & 227$^\dagger$ & 9/5 \\
    \task{fold\_clothes\_random}        &  500 &  6 & 10 &  9    & 55 & 99            & 3            &   5 &   5           &  19           & 20/20 \\
    \task{hang\_mugs\_random}           &  800 &  2 &  1 &  0    & 39 & 100$^\dagger$ & 6$^\dagger$  & 114 & 148$^\dagger$ & 147$^\dagger$ & 2/4 \\
    \task{insert\_key}                  &  300 &  0 &  0 &  0    & 28 & 97            & 3            &  33 &  57           &  76           & 20/20 \\
    \task{insert\_tubes}                &  500 & 12 & 11 &  6    & 38 & 95            & 5            &  40 &  57           & 116           & 20/19 \\
    \task{play\_tic\_tac\_toe}          & 1100 & 12 & 16 & 16    & 38 & 90            & 0            &  19 &  50           &  97           & 20/20 \\
    \task{pour\_balls\_into\_vase}      &  600 &  8 & 11 & 12    & 39 & 75            & 1            &  44 &  56           &  74           & 18/18 \\
    \task{solve\_equation}              &  300 & 14 & 19 & 15    & 27 & 97            & 2            &  33 &  35           &  56           & 20/20 \\
    \task{store\_laptop\_and\_headphones} & 800 &  0 &  0 &  1    & 22 & 82$^\dagger$  & 1$^\dagger$  &  63 &  68$^\dagger$ & 145$^\dagger$ & 9/5 \\
    \task{store\_tools\_in\_toolbox}    &  900 &  0 &  0 &  0    & 29 & 95$^\dagger$  & 63$^\dagger$ & 140 & 152$^\dagger$ & 141$^\dagger$ & 9/1 \\
    \midrule
    Total & & 65/200 & 80/200 & 71/200 & 35 & 91 & 3 & -- & -- & -- & 147/132 \\
    \bottomrule
  \end{tabular}
\end{table}

On every task, on-10 makes the most re-measure moves of the three libraries; in actions it spends the most on seven
of the ten tasks, and off-10 slightly more on \task{fasten\_screws}, \task{hang\_mugs\_random} and
\task{store\_tools\_in\_toolbox}, where off-10 and on-10 have few such episodes. Of the 14 failed
on-10 episodes of \task{insert\_tubes}, 9 used at least 95\% of the budget, and its successful episodes needed 442
actions on average against 397 (empty) and 392 (off-10).

\begin{figure}[!htbp]
  \centering
  \footnotesize
  \setlength{\tabcolsep}{1pt}
  \begin{tabular}{@{}c cccc@{}}
    & action 120 & action 250 & action 330 & last action \\
    \rotatebox{90}{\makebox[0.17\linewidth]{$\Lempty$}} &
    \includegraphics[width=0.235\linewidth]{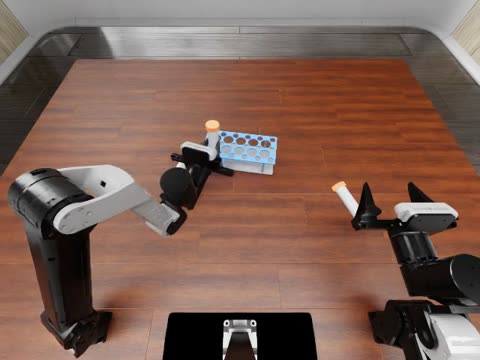} &
    \includegraphics[width=0.235\linewidth]{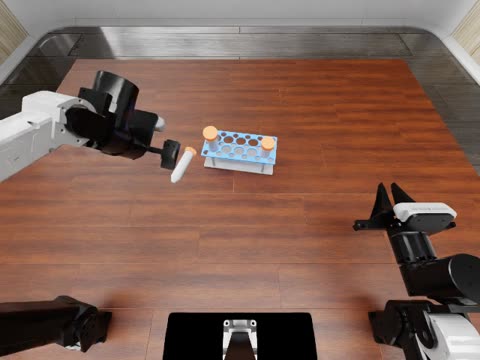} &
    \includegraphics[width=0.235\linewidth]{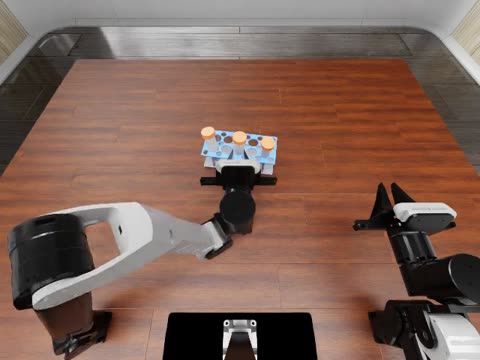} &
    \includegraphics[width=0.235\linewidth]{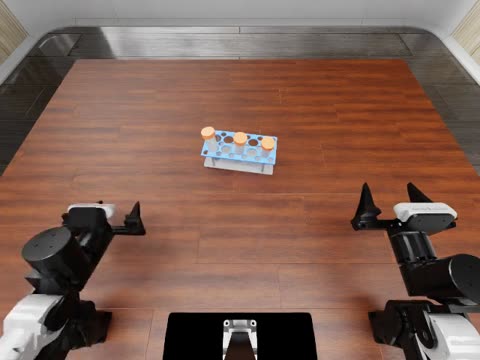} \\
    \rotatebox{90}{\makebox[0.17\linewidth]{off-10}} &
    \includegraphics[width=0.235\linewidth]{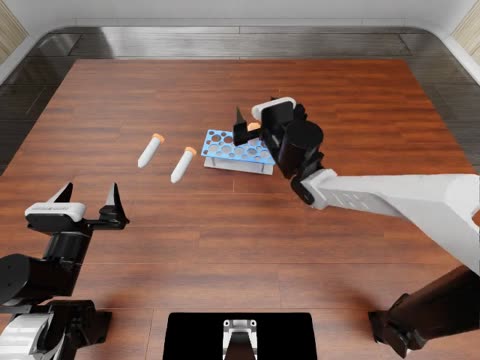} &
    \includegraphics[width=0.235\linewidth]{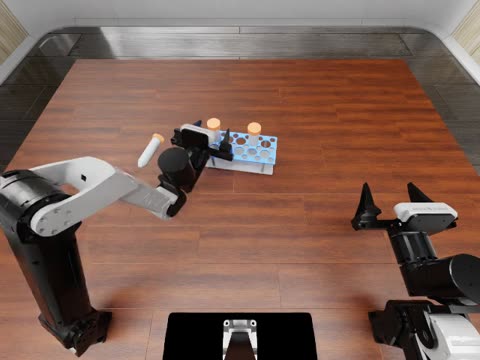} &
    \includegraphics[width=0.235\linewidth]{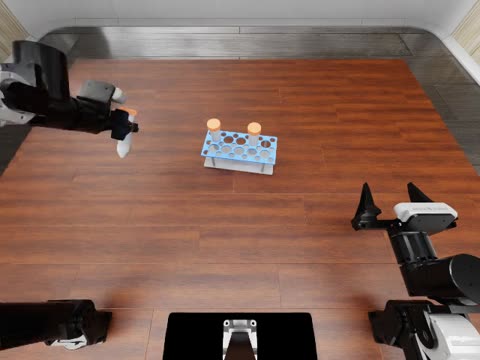} &
    \includegraphics[width=0.235\linewidth]{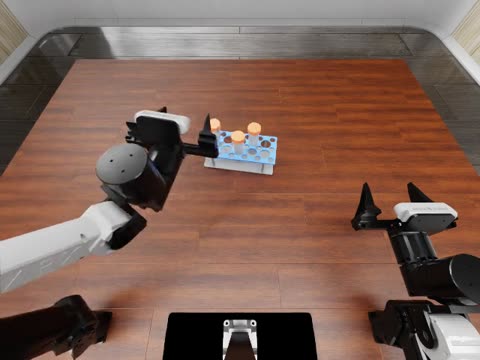} \\
    \rotatebox{90}{\makebox[0.17\linewidth]{on-10}} &
    \includegraphics[width=0.235\linewidth]{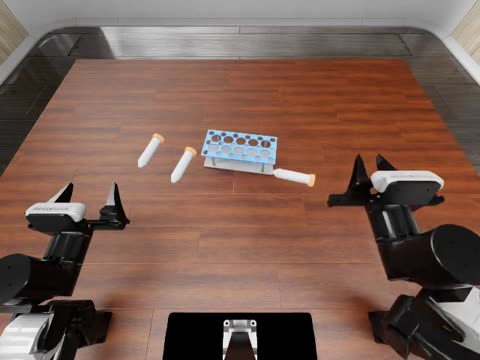} &
    \includegraphics[width=0.235\linewidth]{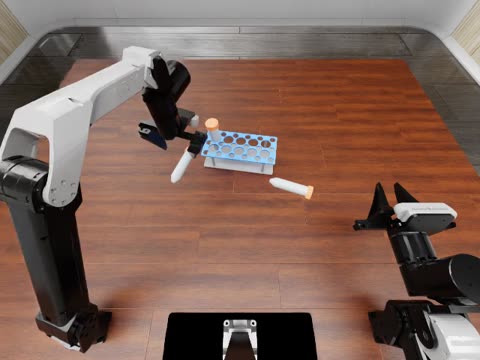} &
    \includegraphics[width=0.235\linewidth]{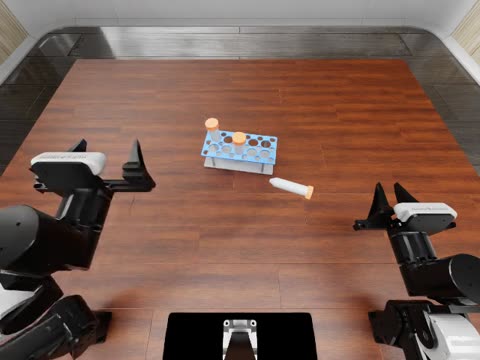} &
    \includegraphics[width=0.235\linewidth]{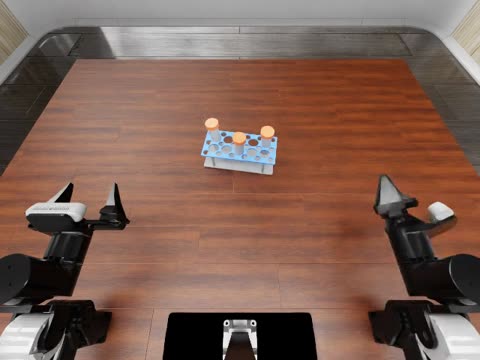} \\
  \end{tabular}

  \vspace{4pt}
  \begin{tabularx}{\linewidth}{@{}l l >{\raggedright\arraybackslash}X@{}}
    \toprule
    Library & Result & Record \\
    \midrule
    $\Lempty$ & success, 352 & No re-measure move: inserts from the head-camera estimates, releasing the tubes at
      actions 123, 213 and 327; judged a success during the return home. \\
    off-10 & success, 388 & One re-measure move (25 actions); releases at actions 115, 249 and 373, 3 blocked
      descents; judged a success as the arm withdraws. \\
    on-10 & fail, 499 & 9 re-measure moves (126 actions) and wrist triangulation (\texttt{triangulate}, as asked by
      \texttt{place-008}). The first tube falls out after its release at action 100; the other two are seated at
      actions 215 and 307; the fallen tube is grasped again at action 370 and seated at action 478. The budget runs out
      at action 499, before the arms are home. \\
    \bottomrule
  \end{tabularx}
  \caption{\textbf{\task{insert\_tubes}, layout 19: re-measuring uses up the budget.} Three tubes go into separate
    holes of the rack within 500 actions. Head camera after the given action, rows the empty library, off-10 and
    on-10; the last column is each episode's last action (352, 388, 499). At action 330 the empty library inserts its
    third tube while on-10's first tube lies on the table; at its last action on-10 has all three tubes in the rack
    but has not withdrawn the arm.}
  \label{fig:case-tubes}
\end{figure}

Two layouts show the difference step by step. On layout 4 of \task{solve\_equation}, a thin disc has to be placed on a
pad within 300 actions: off-10 passes \texttt{OPEN} on every move of an empty gripper and holds the disc at its first
close (success after 125 actions), while on-10 descends half open with \texttt{KEEP}, all five of its closes grasp
nothing, and it ends after 292 actions with the pad empty. Layout 19 of \task{insert\_tubes} is in
\cref{fig:case-tubes}.

\section{Why Cross-benchmark Libraries Help Little}
\label{app:xbench}

\paragraph{Setup.} The libraries off-10 and on-10 of Appendix~\ref{app:case} are translated to the prompt and primitives of
LIBERO-PRO and RoboCasa365; off-rc10 and on-rc10 are distilled from RoboCasa365 tasks disjoint from its ten target
tasks. At the start of an episode the library is placed in the actor's working directory: a skill runs only if the
actor's code imports it, while the memory entries are appended to the prompt. We count the episodes whose code imports
or calls a skill.

\paragraph{Results.} Three reasons explain why neither setting's library gives a clear gain across benchmarks.

\textbf{First, the translated skills are almost never called.} Even within RoboDojo only \V{XB_USE_R10_OFF}\% (off-10)
and \V{XB_USE_R10_ON}\% (on-10) of the episodes call a skill. None of the 200 RoboCasa365 episodes with off-10 or
on-10 calls one, and on LIBERO-PRO only \V{XB_LB_CALL_OFF} and \V{XB_LB_CALL_ON} of 60 episodes do. The libraries
distilled within RoboCasa365 are called in \V{XB_RC_CALL_RC} of 200 episodes, mostly their perception functions for
two-view localization and joint margins. Across benchmarks, the memory entries appended to the prompt are the only part
of a translated library the actor uses.

\textbf{Second, what the translated libraries hold is not what the targets need.} They cover grasping from above,
placing and withdrawing at a table; translation removed the content about two arms, hand-overs and the table geometry,
and found no source experience for moving the base. Most of the ten RoboCasa365 target tasks open doors, turn knobs,
pull drawers or reach into racks, and need the base to move. The libraries distilled within RoboCasa365 cover exactly
these: moving the base in front of a door, swinging a hinged part about its axis, approaching under an overhang, and
recovering when the arm is near its joint limits. The perception experience does not carry either: the millimetre-level
precision learned in RoboDojo does not hold on RoboCasa365, where the measured height of an object top is off by
2--8\,cm.

\textbf{Third, LIBERO-PRO leaves little room.} With the aligned primitives, the empty library already solves
\V{XB_LB_ALL_V0} of 200 episodes on the four swap suites, and only six of the 40 tasks are not solved on every seed. On
four of these six every library solves 8--10 of 10 seeds; the differences come from two tasks, putting a book into the
back compartment of a caddy (empty, off-10, on-10: \V{XB_LB_T5} of 10) and putting a mug into the microwave and closing
it (\V{XB_LB_T9}). The same library solves different numbers on the two halves of the seeds (empty \V{XB_LB_HALF_V0},
off-10 \V{XB_LB_HALF_OFF}, on-10 \V{XB_LB_HALF_ON}, of 30 each), a spread larger than the gain of any library.

\paragraph{Interpretation.} Transfer helps where source and target share the bottleneck. Within RoboDojo the robot, the
table and perception stay the same, and off-10 gains on target tasks. Across benchmarks, the general part of the
experience (grasping, placing, finishing) is what the target's prompt and the actor already handle, and what the
target lacks, moving the base and operating articulated parts, has no counterpart in the source. A library distilled
on the target benchmark, even from other tasks, covers that bottleneck, is called in most episodes, and helps. Two
caveats apply: a skill that is never called does not mean the library has no effect, since its memory entries are
always in the prompt; and we evaluated one way of translating a library, not variants such as skills without memory.

\FloatBarrier
\section{Cross-benchmark Primitive Alignment}
\label{app:align}
With the aligned primitives, the empty library solves 41, 47, 50 and 50 of 50 episodes on the four LIBERO-PRO swap
suites (10, goal, object, spatial), against 36, 45, 44 and 39 for the published PyRUA-Lean~\citep{pyrualean2026}; on
RoboCasa365 it solves \V{RCA_ATOMIC}/90 atomic and \V{RCC}/160 composite episodes, against \V{RCA_ATOMIC_PAPER}/90 and
\V{RCC_PAPER}/160 published. \Cref{tab:align} summarises the comparison.

\begin{table}[!htbp]
  \centering
  \caption{\textbf{Primitive alignment.} Successes of the empty library with the primitives aligned to RoboDojo,
    against the published no-VLA PyRUA-Lean code agent on the same tasks and seeds~\citep{pyrualean2026}.}
  \label{tab:align}
  \small
  \begin{tabular}{@{}l l c c@{}}
    \toprule
    Benchmark & Suite & Aligned & Published \\
    \midrule
    LIBERO-PRO & \texttt{libero\_10\_swap} & 41/50 & 36/50 \\
    & \texttt{libero\_goal\_swap} & 47/50 & 45/50 \\
    & \texttt{libero\_object\_swap} & 50/50 & 44/50 \\
    & \texttt{libero\_spatial\_swap} & 50/50 & 39/50 \\
    RoboCasa365 & atomic & \V{RCA_ATOMIC}/90 & \V{RCA_ATOMIC_PAPER}/90 \\
    & composite & \V{RCC}/160 & \V{RCC_PAPER}/160 \\
    \bottomrule
  \end{tabular}
\end{table}

On LIBERO-PRO, \V{LB_FULL} of the \V{LB_TASKS} tasks are solved on all five seeds with the empty library; the other
six are the tasks of \cref{sec:exp-bench}.

\FloatBarrier

\end{document}